\documentclass[11pt]{article}

\usepackage{arxiv}

\usepackage{amsmath,amsfonts,bm}
\usepackage{booktabs}          
\usepackage{graphicx}          
\usepackage{subcaption}        

\def\eqref#1{Eq.~\ref{#1}}   
\newcommand\numberthis{\addtocounter{equation}{1}\tag{\theequation}}       

\def\eps{{\epsilon}}

\def\vtheta{{\bm{\theta}}}
\def\vTheta{{\bm{\Theta}}}
\def\vbeta{{\bm{\beta}}}

\def\vpsi{{\bm{\psi}}}
\def\valpha{{\bm{\alpha}}}
\def\va{{\bm{a}}}
\def\vb{{\bm{b}}}

\def\vd{{\bm{d}}}

\def\vm{{\bm{m}}}
\def\vn{{\bm{n}}}

\def\vx{{\bm{x}}}
\def\vy{{\bm{y}}}
\def\vz{{\bm{z}}}

\def\mA{{\bm{A}}}
\def\mB{{\bm{B}}}

\def\mI{{\bm{I}}}

\def\mSigma{{\bm{\Sigma}}}

\newcommand{\E}{\mathbb{E}}
\newcommand{\Var}{\mathrm{Var}}

\newcommand{\simiid}{\overset{\mathrm{iid}}{\sim}}
\newcommand{\pd}{{\color{black} p_d}}
\newcommand{\pn}{{\color{black} p_n}}
\newcommand{\fisherscore}[1][\,\cdot\,]{{\color{black} 
\nabla_{\vbeta} F(#1; \vbeta^*)}}
\newcommand{\Ifisher}{{\color{black} \bm{\mathcal{J}}}}

\newcommand{\R}{\mathbb{R}}
\newcommand{\sigmoid}{\sigma}

\DeclareMathOperator*{\argmin}{\arg\!\min}

\usepackage{amsthm}

\newtheoremstyle{mythmstyle} 
{\topsep}    
{\topsep}    
{\itshape}   
{0pt}        
{\bfseries}  
{}           
{ }          
{}           
\theoremstyle{mythmstyle}
\newtheorem{theorem}{Theorem}
\newtheorem{conjecture}{Conjecture}

\newcommand{\omar}[1]{\textcolor{black}{#1}}

\newcommand{\aapo}[1]{\textcolor{black}{#1}} 

\begin{document}

\title{Optimizing the Noise in Self-Supervised Learning: \\ from Importance Sampling to Noise-Contrastive Estimation}
\date{}

\author{
    Omar Chehab$^{1}$
    \footnote{
    Corresponding author \href{mailto:l-emir-omar.chehab@inria.fr}{l-emir-omar.chehab@inria.fr}},
    {Alexandre Gramfort}$^1$,
    {Aapo Hyvarinen}$^{2}$ \\ [.8em]
    $^1${Universit\'e Paris-Saclay, Inria, CEA} \\
    $^2$University of Helsinki \\
}

\maketitle

\begin{abstract}
    
Self-supervised learning is an increasingly popular approach to unsupervised learning, achieving state-of-the-art results. A prevalent approach consists in contrasting data points and noise points within a classification task: this requires a good noise distribution which is notoriously hard to specify. While a comprehensive theory is missing, it is widely assumed that the optimal noise distribution should in practice be made equal to the data distribution, as in Generative Adversarial Networks (GANs).
We here empirically and theoretically challenge this assumption. 
We turn to Noise-Contrastive Estimation (NCE) which grounds this self-supervised task as an estimation problem of an energy-based model of the data. This ties the optimality of the noise distribution to the sample efficiency of the estimator, 
which is rigorously defined as its asymptotic variance, or mean-squared error.
In the special case where the normalization constant only is unknown, we show that NCE recovers a family of Importance Sampling estimators for which the optimal noise is indeed equal to the data distribution. 
However, in the general case where the energy is also unknown, we prove that the optimal noise density is the data density multiplied by a correction term based on the Fisher score. In particular, the optimal noise distribution is different from the data distribution, and is even from a different family.
Nevertheless, we soberly conclude that the optimal noise 
may be hard to sample from, and the gain in efficiency can be modest compared to choosing the noise distribution equal to the data's.

\end{abstract}


\section{Introduction}
\label{sec:intro}

Self-supervised learning is a powerful and increasingly successful approach for unsupervised learning. It consists in reformulating an unsupervised task (\textit{e.g.}, learning features from data) as a supervised task (\textit{e.g.}, classification). A fundamental instance of this is binary classification~\citep{chen2020simclr,vanoord2018cpc,hyvarinen2017pcl}, for example between data points and noise points~\citep{gutmann2012nce}.
The supervised task is a ``pretext'' task that is easy to understand and simple to implement, while of little interest in itself. The actual utility of such learning might be, for example, that it recovers interesting features in the hidden layers of a neural network.

While self-supervised learning has demonstrated broad empirical success in applications to natural language processing~\citep{mikolov2013nce,mnih2012condnce}, vision~\citep{vanoord2018cpc,chen2020simclr}, or neuroscience~\citep{banville2020ssleeg}, a key question is how to \textit{design} the pretext task. In the case of binary classification, some of the essential design choices are: which noise samples should be contrasted with the data samples? in what proportion? and using which classification loss? 
This of course depends on the end goal. Often, the classifier is used to learn features that are evaluated by classifying another dataset of interest, i.e. ``downstream". This intimately ties the design of the classification task to the dataset used for evaluation~\citep{saunshi2019ssltheory,tsai2020sslmutualinfo}.
Unfortunately, such datasets vary across applications 
and therefore do not easily lead to any definitive conclusions regarding the optimal choice of the pretext task.

One approach that provides rigorous and general answers to the question of optimal design is given by statistical estimation theory, especially in the case of Noise-Contrastive Estimation (NCE)~\citep{gutmann2012nce}. Using binary classification as a pretext text, NCE effectively estimates an unnormalized (or energy-based) statistical model of the data distribution.
In this setting, the sample efficiency of the data estimator can be rigorously measured as the asymptotic MSE (or variance) of the model parameters.
Using such an analysis one can study the optimality of the design choices (\textit{e.g.} noise distribution) of the pretext task, thus complementing some initial results already given by \citet{gutmann2012nce}.

Learning an unnormalized (or energy-based) model of a data distribution is actually a well-known statistical problem that has seen renewed interest as it is brought to scale with deep learning.
 For example, classical Monte-Carlo approaches such as Importance Sampling rely on a noise distribution to estimate the normalization constant of the model. In this case, characterizing the optimal noise based on asymptotic variance is well-understood~\citep{owenmontecarlobook} and could likely provide insight into choosing the optimal noise for self-supervised learning, in particular NCE.
 
Conventional wisdom in deep learning, prompted by the literature on generative adversarial networks (GANs) \citep{goodfellow2014gan, gao2020fce}, is to set the distribution of noise to be equal to the data's. The underlying intuition 
is that the task of discriminating data from noise is hardest, and therefore most ``rewarding'', when noise and data are virtually indistinguishable by a classifier. 
However, while the above assumption (optimal noise equals data) has been supported by numerous empirical successes,
it is not clear whether such a choice of noise achieves optimality from a statistical estimation viewpoint.

In this work, we show how statistical estimation theory provides a principled approach for designing the pretext task of NCE. 
We start by proving a connection between Importance Sampling and NCE in the case where the parameters are known and only the normalization constant is estimated. We show that NCE then recovers a well-known family of Importance Sampling estimators, for which the noise distribution is optimally equal to the data's. 
Next we show how NCE generalizes Importance Sampling in the case of estimating an energy-based model in full, i.e. both the normalization and the parameters in the energy. We analytically provide the optimal noise for NCE in the two aforementioned limits.
The latter general analysis leads to the following claims that challenge conventional wisdom:
\begin{itemize}
    \item An optimal noise distribution is not the data distribution; in fact, it is of a very different family than the model family.
    \item There can be many equivalent optimal noise distributions: they have in common that they emphasize parts of the data space where the learnt classifier is most sensitive to its parameterization
    \item The optimal noise proportion is generally not 50\%; the optimal noise-data ratio is not one.
\end{itemize}
However, we soberly claim that while the optimal noise distribution can be very different to data distribution, the sample efficiency is not very different, and sampling from the optimal noise distribution can be very difficult.

This paper\footnote{Preliminary results were presented in \citet{chehab2022nceoptimal}.}
is organized as follows. First, in Section~\ref{sec:background}, we present NCE and related methods, as well as the theoretical framework of asymptotic variance or MSE that we use to optimize the noise distribution.
Our main theoretical results are in Sections~\ref{sec:theoretical_results_partition_model} and~\ref{sec:theoretical_results_general}. In Section~\ref{sec:theoretical_results_partition_model}, we consider NCE in a simple case where the normalization constant only is estimated.
In Section~\ref{sec:theoretical_results_general}, we consider the case where both the normalization constant and the model (energy) parameters are estimated. 
Section~\ref{sec:numerical_results} empirically verifies our predicted optimal noise distribution. 
Finally we discuss current limitations and possible extensions of this work and the NCE framework in Section~\ref{sec:discussion}, before concluding in Section~\ref{ssec:conclusion}.

\section{Background on Noise-Contrastive Estimation}
\label{sec:background}

We start by describing two learning tasks: classification (as a pretext task for self-supervised learning) and model estimation. Then, we point out how Noise-Contrastive Estimation (NCE) connects both tasks. This enables well-established estimation theory to be used to inform the optimal design for the pretext task in self-supervised learning.

\subsection{Designing a Classification Task}
\label{ssec:ssl_classification}

We start by considering the problem of classification which will form the basis for self-supervised learning in later sections.
Arguably, the simplest classification task one could imagine for self-supervised learning is to distinguish a data sample
$(\vx_i)_{i  \in [1, T_d]} \simiid \pd$ 
from a noise sample $(\vx_i)_{i  \in [1, T_n]} \simiid \pn$ using a
classifier $F(\vx)$, as proposed by \citet{gutmann2012nce}. This defines a binary task where $Y=1$ is the data label and $Y=0$ is the noise label. The statistically optimal decision function is given by the log decision barrier (or logit) between the two classes
\begin{align}
    F^*(\vx) 
    &= 
    \log \frac{P(Y=1|\vx)}{P(Y=0|\vx)}
    =
    \log \frac{P(\vx | Y=1) P(Y=1)}{P(\vx|Y=0) P(Y=0)}
    =
    \log \frac{\pd(\vx)}{\nu \pn(\vx)}
    \label{eq:optimal_classifier}
\end{align} 
where $\nu=P(Y=0)/P(Y=1)$ denotes the noise-data ratio. In other words, estimating the optimal classifier $F^*$ means learning 
a density ratio $\pd/\pn$ \citep{gutmann2012nce, mohamed2016implicitgen}. 

While many choices of classification losses are possible, a number of those can be written as a divergence between the optimal classifier $F^*(\vx)$ and the model classifier $F(\vx)$, which are often expressed as functions of their density ratios  $r^* = \frac{\pd}{\nu \pn}$ and $r$ \citep{gutmann2011bregman,menon2016nce,uehara2018nce}.
Bregman divergences offer a class of divergences that are very practical for such density ratios.
This leads to the Bregman classification loss as defined in~\citep[Eq. 13]{gutmann2011bregman}
\begin{align}
    \mathcal{L}(r)
    &=
    D_{\phi}(r^*, r) 
    =
    \nu \E_{\vx \sim \pn}
    \bigg[
    S_0
    \big( r(\vx) \big)
    \bigg]
    -
    \E_{\vx \sim \pd}
    \bigg[
    S_1
    \big ( r(\vx) \big)
    \bigg]
    \label{eq:gnce_bregman_ratio}
    + \mathrm{constant}
\end{align}
where $(S_0, S_1)$ are nonlinearities used for notational convenience following~\citet{pihlaja2010nce,gutmann2011bregman}; but which actually depend on a single, convex function $\phi$ as
\begin{alignat}{2}
    S_0(x) = -\phi(x) + \phi'(x)x  
    \hspace{2em}
    S_1(x) = \phi'(x)
\end{alignat}
or conversely
\begin{align}
    \phi(x) &= S_1(x) x - S_0(x)
    \enspace .
\end{align}
Different choices of $\phi$ lead to different classification losses, which we name following the terminology of~\citep[Section 2.2]{uehara2018nce}. For example, $\phi(x) = x\log x - (1+x)\log(\frac{1+x}{2})$ recovers the Jensen Shannon (JS) classification loss (more commonly known as the logistic loss). Another choice, $\phi(x) = (1- \sqrt{x})^2$, recovers the squared Hellinger ($H^2$) classification loss (more commonly known as the exponential loss). Two additional choices of $\phi$ will be of interest to us in this manuscript: $\phi(x) = x \log x$ defining the Kullback-Leibler (KL) classification loss, and $\phi(x) = -\log(x)$ defining the reverse-KL classification loss. Further details on terminology and usage of the Bregman loss for classification is provided in Appendix~\ref{app:ssec:nce_bregman_formulation}. \\ \\
So far, the \textit{design} of the classification task relies on four choices:
\begin{itemize}
    \item the classification ``contrast" or noise distribution $p_n$ 
    \item the classification imbalance, given by the noise-data ratio $\nu = T_n/T_d$ 
    \item the classification loss (\textit{e.g.} logistic, exponential)
    indexed by a convex function $\phi$
    \item the computational budget, corresponding here to the total number of observations $T= T_d + T_n$
\end{itemize}
These four hyperparameters will influence the estimation of the classifier. 

\subsection{Parametric Statistical Models}
\label{ssec:parametric_models}

Another well-known learning problem is estimation of parameters in a statistical model. In this subsection we define that problem; we will show in the next subsection how it is related to classification.

We consider parametric families for the data model $p(\cdot; \vtheta)$, with $\vtheta \in \R^d$. As is typical in estimation theory, we assume that the actual data distribution $p_d$ belongs to the parametric family, i.e.\ the model is well-specified. This means that there exists $\vtheta^* \in \R^d$ such that: 
\begin{align}
    \pd(\vx)
    = p(\vx; \vtheta^*)
\end{align}

\paragraph{A normalized model} Classical estimation theory typically considers a normalized model obtained as
\begin{align}
    p(\vx; \vtheta)
    &= \frac{\tilde{p}(\vx; \vtheta)}{Z(\vtheta)}
    \quad
    \text{where } Z(\vtheta) = \int \tilde{p}(\vx; \vtheta) d\vx
    \label{eq:normalized_model} \enspace .
\end{align}
The normalizing constant $Z(\vtheta)$ is entirely determined by the unnormalized density $\tilde{p}$, which can be arbitrarily chosen as long as it is non-negative and integrable for all $\vtheta$. There are families of distributions where the normalization term can be explicitly evaluated, either because the integral is simple enough to be evaluated in closed form (\textit{e.g.} for the Gaussian family) or because it is explicitly expressed in terms of a latent variable by the change of variables formula (\textit{e.g.} for a normalizing flow family \cite{dinh2016realnvp}). Yet, having such a tractable way to evaluate the normalization constant limits the expressiveness of the data models one can consider~\citep{gao2020fce}. 

In most cases, $Z(\vtheta)$ is an intractable integral for which numerical integration such as quadrature can be considered. Yet, this becomes prohibitive with a complexity that is exponential in the dimension of $\vx$.
This motivated using probabilistic integration methods such as Importance Sampling, where independent samples drawn from an auxiliary distribution are used to approximate the integral. The exponential cost in the dimension of $\vx$ is now shifted to the choice of the auxiliary distribution~\citep[Example 9.1]{owenmontecarlobook}, which is crucial.
Alternatively, we may consider unnormalized models which are discussed next.

\paragraph{A unnormalized  (a.k.a. Energy-Based) model}
Our main interest in this paper is to estimate the parameter $\vtheta$ without explicitly evaluating the normalization constant $Z(\vtheta)$ via a multidimensional integral.
This can be done either by methods that \textit{``ignore"} the normalization constant (\textit{e.g.} Score-Matching~\citep{hyvarinen2005scorematching}, Contrastive Divergence~\citep{hinton2002contrastivedivergence}) or methods that estimate it \textit{jointly} with the parameters 
--- the next subsection will show how NCE was primarily conceived for this purpose. Thus, we introduce another free parameter $Z$ where the dependency on $\vtheta$ is dropped:
\begin{align}
    p(\vx; (\vtheta, Z))
    &= \frac{\tilde{p}(\vx; \vtheta)}{Z} \enspace .
\end{align}
 We use the term ``estimation" permissively, as $Z$ is \textit{not} considered a parameter in classical statistical theory, yet for unnormalized models it is nonetheless approximated from a sample.
In practice, we often use the log normalization constant instead, $c = \log Z$. What motivates this parametrization by $c$ is that it is conveniently unconstrained (need not be positive) and its Fisher score is constant $\nabla_{c} \log p(\vx; (\theta, c)) = -1$, which simplifies statistical analyses~\citep[Appendix B.1.]{gutmann2012nce}. 

\paragraph{Computing the normalization constant} An interesting special case of unnormalized models is when the energy is known, and only the normalization constant $Z$ is not:
\begin{align}
    p(\vx; Z)
    &=  \frac{\tilde{p}(\vx; \vtheta^*)}{Z}
    =  \frac{f(\vx)}{Z}
    \label{eq:unnormalized_partition_model}
    \enspace ,
\end{align}
using $f(\vx)$ as a shorthand notation. There remains to 
``estimate" the normalization constant.

Whenever we state general results which hold for any choice of parameterization, we will use $p( \cdot; \vbeta)$ where $\vbeta$ is a generic parameter, often shortened to $p_\vbeta(\cdot)$. For a normalized model, $\vbeta = \{ \vtheta \}$ while for an unnormalized model $\vbeta = \{ \vtheta, c \}$. 

\subsection{Noise-Contrastive Estimation and its variants}
\label{ssec:nce}

Noise-Contrastive Estimation (NCE) connects the task of \textit{classification}, which was above considered to happen by estimating the density ratio $\pd / \pn$, with the task of \textit{model estimation}, which is about fitting the data density $\pd$. To do so, the noise distribution $\pn$ is specified by the user and the classifier is parameterized like the optimal classifier in~\eqref{eq:optimal_classifier}
\begin{align}
    F(\vx; \vbeta) 
    = 
    \log 
    \frac{p_\vbeta(\vx)}{\nu\pn(\vx)}
    =
    \log p_\vbeta(\vx)
    - 
    \log (\nu\pn(\vx))
    \label{eq:model_classifier}
\end{align}
only replacing the data distribution $\pd$ with the model distribution $p_{\vbeta}$. This way, learning the classifier is equivalent to estimating a parametric model $p_\vbeta$ of the data distribution. 
Note that while the classifier $F$ and log-likelihood are different as in~\eqref{eq:model_classifier}, their parameter-gradients are equal
\begin{align} \label{eq:Flogp}
    \nabla_{\vbeta} \log p_\vbeta(\vx)  
    =
    \nabla_{\vbeta} F(\vx; \vbeta) 
    \enspace 
\end{align}
and are known as the Fisher score for ordinary statistical models. 

Importantly, NCE can estimate a wider class of models than its counterparts: $p_\vbeta$ can be an unnormalized parameterization of the data distribution (See definitions in Section~\ref{ssec:model_normalized}). 
In fact, Bregman divergences can directly handle unnormalized distributions $p_\vbeta$; there is no assumption on normalization done in that theory.
The normalization constant can be estimated as an additional free parameter, in stark contrast to Maximum-Likelihood Estimation (MLE). A general form of 
the NCE estimator $\hat{\vbeta}$ is thus obtained by minimizing a Bregman classification loss
\begin{align}
    \mathcal{L}(\vbeta)
    &=
    \nu \E_{\vx \sim \pn}
    \bigg[
    S_0
    \bigg(
    \frac{p_\vbeta}{\nu p_n}(\vx)
    \bigg)
    \bigg]
    -
    \E_{\vx \sim \pd}
    \bigg[
    S_1
    \bigg(
    \frac{p_\vbeta}{\nu p_n}(\vx)
    \bigg)
    \bigg]
    \enspace .
    \label{eq:gnce_bregman}
\end{align}
 As the minimizer, the NCE estimator is thus a function of the samples $\vx_{0:T}$ 
and the hyperparameters $(\phi, \nu, \pn, T)$ of the classification task. 
A few special cases are listed in Table~\ref{table:classification_loss_generic}.

\paragraph{Original NCE}
For later developments, it is particularly useful to write the estimator as the solution when the gradient of the classification loss is zero.
In the case of the JS classification loss originally proposed for NCE~\citep{gutmann2012nce} and recalled in Table~\ref{table:classification_loss_generic}, this is written as
\begin{align}
    0
    &=
    \nabla_{\vbeta} 
    \mathcal{L}(\vbeta)
    =
    \nu \E_{\vx \sim \pn}
    \bigg[
    w(\vx, 0)
    \nabla_{\vbeta} 
    \log p_{\vbeta}(\vx)
    \bigg]
    -
    \E_{\vx \sim \pd}
    \bigg[
    w(\vx, 1)
    \nabla_{\vbeta} 
    \log p_{\vbeta}(\vx)
    \bigg]    
    \enspace ,
    \label{eq:nce_implicit_logistic}
\end{align}
where the reweighting function
\begin{align}
    w(\vx, y) 
    =
    \begin{cases}
        p_{\vbeta}(Y=0|\vx) := \sigmoid( \log ( \nu \pn(\vx) /  p(\vx, \vbeta) ) ) & \text{if } y=1
        \\
        p_{\vbeta}(Y=1|\vx) := \sigmoid ( \log ( p(\vx, \vbeta) / \nu \pn(\vx) ) )  & \text{if } y=0
    \end{cases}
    \label{eq:nce_original_reweighting}
\end{align}
reweights points by their ``misclassification score" in the range of $[0, 1]$, as $\sigmoid$ denotes the sigmoidal function. It effectively selects misclassified points and ``zeroes-out" perfectly classified points.
To gain some intuition, suppose the model is normalized, so that the Fisher score $\nabla_{\vbeta} \log p_{\vbeta}(\vx)$ is in fact the gradient of the log-likelihood. With each gradient step $\vbeta \leftarrow \vbeta - \mu \nabla_{\vbeta} \mathcal{L}(\vbeta)$, the log-likelihood is increased for the reweighted data sample (second term) and decreased for the reweighted noise sample (first term). The contribution of each data (or noise) point is reweighted by the misclassification error, and this process terminates when it reaches the NCE estimator $\hat{\vbeta}$. 
The following generalizations of NCE have consisted in generalizing (relaxing) this implicit definition of the estimator, sometimes at the risk of departing from the classification framework.

\paragraph{Relaxing the reweighting}
Choosing a Bregman classification loss defined via $\phi$ extends the class of reweighting functions to
\begin{align}
    w(\vx, y) 
    =
    \begin{cases}
        w(\vx)
        & \text{if } y=1
        \\
        w(\vx) \frac{p_{\vbeta}}{\nu \pn}(\vx)
        & \text{if } y=0
    \end{cases}
    \label{eq:bregman_reweighting}
    \enspace .
\end{align}
where
\begin{align}
    w(\vx)
    =
    \frac{p_{\vbeta}}{\nu \pn}(\vx) 
    \phi^{''} \big(
    \frac{p_{\vbeta}}{\nu \pn}(\vx)
    \big) 
    \enspace .
    \label{eq:reweighting_function}
\end{align}
As in~\eqref{eq:nce_implicit_logistic}, minimizing the loss still increases the log-likelihood of the reweighted data sample and decreases the log-likelihood of the reweighted noise sample; the contribution of each point is reweighted by a function that now depends on the classification loss identified by $\phi$, following \eqref{eq:bregman_reweighting}. 
For example, the JS classification loss yields $\phi^{''}(x) = \frac{1}{x(1 + x)}$ and recovers the reweighting function of \eqref{eq:nce_original_reweighting}.
With this generalization, equation~\eqref{eq:nce_implicit_logistic} can be re-arranged as
\begin{align}
    0
    &=
    \nabla_{\vbeta} 
    \mathcal{L}(\vbeta)
    =
    \nu \E_{\vx \sim \pn}
    [
    p_{\vbeta}(\vx)
    \,
    \valpha(\vx)
    ]
    -
    \E_{\vx \sim \pd}
    [
    \pn(\vx)
    \,
    \valpha(\vx)
    ]    
    \enspace ,
    \label{eq:nce_implicit_alpha}
\end{align}
introducing a vector-valued function 
\begin{align}
    \valpha(\vx) 
    &= 
    w(\vx) \frac{\nabla_{\vbeta} \log p_{\vbeta}(\vx)}{\pn(\vx)}
    \label{eq:connection_is_nce}
    \enspace ,
\end{align}
as done in \citep[Equation 3.24]{uehara2018nce} and in~\citep[Equation 22]{xing2022adaptiveIS}, although the latter did so in a similar context without drawing the connection to NCE. This new form will be useful for Section~\ref{sec:theoretical_results_partition_model}. 

It is natural to ask at this point whether \eqref{eq:nce_implicit_logistic} still provides a valid estimator when the contribution of the data and noise samples is reweighed by a \textit{general} function $w(\vx, y)$ that is \textit{not} derived from a classification loss. 
The answer is yes:  \citet[Definition 3]{uehara2018nce} propose to relax NCE using
a more general reweighting function $w(\vx)$,
no longer constrained by its definition in \eqref{eq:bregman_reweighting}.

\begin{table}[!h]
  \caption{
  Binary classification losses for different choices of convex function $\phi$. 
  This is based on Table 1 in~\citep{pihlaja2010nce}. 
  }
  \centering
  \begin{tabular}{llll}
    \toprule
    Name & Hyperparameter & Classification Loss \\
    {} & $\phi(x)$ & $\mathcal{L}(p_\vbeta; \phi)$ \\
    \midrule
    revKL & $-\log(x)$ & $- \E_{d} \frac{\nu \pn}{p_\vbeta} -\nu \E_{n} \log p_\vbeta$ \\
    \midrule
    JS & $x\log x - (1+x)\log(\frac{1+x}{2})$ & $\E_{d} \log \sigmoid \left( \log \frac{p_\vbeta}{\nu \pn} \right) + \nu \E_{n} \log \sigmoid \left( \log \frac{\nu \pn}{p_\vbeta} \right)$ \\
    \midrule
    KL & $x \log x$ & $\E_{d}  \log p_\vbeta - \nu \E_{n}\frac{p_\vbeta}{\nu \pn}$ \\
    \midrule
    $H^2$ & $(1 - \sqrt{x})^2$ & $\E_d \sqrt{\frac{\nu \pn}{p_\vbeta}} + \nu \E_n \sqrt{\frac{p_\vbeta}{\nu \pn}}$ \\
    \bottomrule
  \end{tabular}
  \label{table:classification_loss_generic}
\end{table}    

For the rest of the manuscript, we will use ``NCE" to refer to the original estimator obtained with the logistic classification loss, and use ``NCE family" to refer to estimators obtained with a Bregman classification loss.

\subsection{Measuring the Estimation Error}
\label{ssec:parametric_estimation_error}

The parametric estimation error, as defined in classical statistical theory, is our focus in this paper. For this reason, we assume that it dominates all other errors, related to optimization or model mis-specification~\citep[Introduction to Statistical Learning Theory, Section 2.2]{bousquet2004stattheory}; the latter does not actually occur by assumption of a well-specified model. This is equivalent to saying that the generalization error of the estimator in the framework of Empirical Risk Minimization (ERM) reduces to the estimation error from estimating $\beta^*$ using a finite sample. 

\paragraph{Estimation Error (parametric)}
The parametric estimation error is measured by the Mean-Squared Error (MSE), which consists of variance and squared bias: 
\begin{align}
    \E_{\hat{\vbeta}} \big[
    \| \hat{\vbeta} - \vbeta^* \|^2
    \big]
    = 
    \mathrm{tr} (
    \Var_{\hat{\vbeta}}(\hat{\vbeta})
    )
    + 
    \|
    \mathrm{bias}_{\hat{\vbeta}}(\hat{\vbeta}, \vbeta^*)
    \|^{2} 
    \enspace .
\end{align}
It can mainly be analyzed in the asymptotic regime, with the data sample size $T_d$ being very large. For (asymptotically) unbiased estimators, the 
estimator's statistical performance is in fact completely characterized by its asymptotic variance matrix~\citep[Eq. 5.20]{vandervaart2000asympstats}, classically defined as
\begin{equation} \label{eq:asvar}
    \mSigma 
    = 
    \lim_{T_d\rightarrow \infty} T_d \,
    \Var_{\hat{\vbeta}}(\hat{\vbeta})
    =
    \lim_{T_d\rightarrow \infty} T_d \, \E_{\hat{\vbeta}}[(\hat{\vbeta} - \E_{\vbeta}[\hat{\vbeta}])(\hat{\vbeta} - \mathbb{E}_{\hat{\vbeta}}[\hat{\vbeta}])^\top]
\end{equation} 
where the estimator is computed from a data sample of size $T_d$. The leaves the Mean-Squared Error as
\begin{align}
    \mathrm{MSE} 
    = 
    \frac{1}{T_d} \mathrm{tr}(\mSigma) 
    \enspace .
    \label{eq:mse}
\end{align}
This scalar quantity is the centerpiece of our analysis.

Classical statistical theory tells us that the best attainable $\mathrm{MSE}$ for the energy parameter $\hat{\vtheta}$
(among unbiased estimators) is the Cramer-Rao lower bound. 
The Cramer-Rao bound is asymptotically attained by Maximum-Likelihood Estimation (MLE) for normalized models. This provides a useful baseline, and implies that asymptotically $\mathrm{MSE}_{\mathrm{NCE}} \geq \mathrm{MSE}_{\mathrm{MLE}}$ for normalized models. 
There is no equivalent result for the estimated log-normalization parameter $\hat{c}$ in the case of unnormalized models. 

In a classical statistical framework, the underlying assumption is that the amount of observed data is limited: the data collection is the bottleneck in estimating the parameters. In contrast, we consider here the case where the bottleneck of the estimation is the computation, while data observations are abundant. 
This is the case in many modern machine learning applications: for example, practically infinite amounts of image or audio data can be collected from the internet. The computation can be taken proportional to the sample size, data and noise points together, which we denote by $T = T_d + T_n$. Still, the same asymptotic analysis framework can be used.

Building on~\citep[Equation 8]{pihlaja2010nce},
we can write $\mathrm{MSE}_{\mathrm{NCE}}$ as a function of $T$ (not $T_d$) to enforce a finite computational budget, giving
\begin{align}
    \mSigma 
    &= 
    \mI_{w}^{-1} \big(
    \mI_{v} 
    -
    (1 + \frac{1}{\nu})
    \vm_{w} \vm_{w}^\top
    \big)
    \mI_{w}^{-1}   
    \\
    \mathrm{MSE}_{\hat{\vbeta}}(p_n, \nu, \phi, T) 
    &=
    \frac{\nu + 1}{T} \mathrm{tr}
    (
    \mSigma
    )
    \label{eq:gnce_asymp_var}
\end{align}
where $\vm_w(\vbeta^*)$, $\mI_w(\vbeta^*)$ and $\mI_v(\vbeta^*)$ are the reweighted mean and covariances of a generalized form of the (Fisher) score vector $\fisherscore[\vx] = \nabla_{\vbeta} \log p_{\vbeta }(\vx; \vbeta^*)$, which may now include the derivative with respect to the normalization constant:
\begin{align}
    \vm_w(\vbeta^*) 
    & = 
    \E_{\vx \sim p_d}
    \big[
    w(\vx)
    \fisherscore[\vx]
    \big]
    \\
    \mI_w(\vbeta^*) 
    & = 
    \E_{\vx \sim p_d}
    \big[
    w(\vx)
    \fisherscore[\vx]
    \,
    \fisherscore[\vx]^\top 
    \big]
    \\
    \mI_v(\vbeta^*) & = 
    \E_{\vx \sim p_d}
    \big[
    v(\vx)
    \fisherscore[\vx]
    \,
    \fisherscore[\vx]^\top 
    \big]
    \label{eq:gnce_asymp_integrals}
\end{align}
where the reweighting of data points is by $w(\vx)$ defined in~\eqref{eq:reweighting_function} and by $v(\vx) = w(\vx)^2 p(Y=0 | \vx)^{-1}$, which are all evaluated at the true parameter value $\vbeta^*$. Note that for normalized models (and without the reweighting term), the score's actual  mean is null and its covariance is the Fisher information matrix, written as $\Ifisher$ for the rest of the paper.

\paragraph{Estimation Error (non-parametric)}
So far, we have considered the parametric estimation error of NCE, that is, the error in estimating the data \textit{parameter}. However, sometimes estimating the parameter is only a means for estimating the data \textit{distribution} --- not an end in itself. We therefore also consider the corresponding estimation error induced by the NCE estimator $\hat{\vbeta}$ in distribution space. The non-parametric estimation error is equal to (see Appendix~\ref{app:sec:estimation_error_gnce}):
\begin{align}
    \mathbb{E}\big[ \mathcal{D}_{\mathrm{KL}}(p_d, p_{\hat{\vbeta}}) \big]
    =
    \frac{\nu + 1}{2T}\mathrm{tr}(\mSigma(\vbeta^*) \mI(\vbeta^*))
    \enspace ,
    \label{eq:avgklerror}
\end{align}
where $\mI$ is the (unweighted) covariance of the generalized (Fisher) score vector
\begin{align}
    \mI(\vbeta^*) & = 
    \E_{\vx \sim p_d}
    \big[
    \fisherscore[\vx]
    \,
    \fisherscore[\vx]^\top 
    \big]
    \enspace .
    \label{eq:unweighted_covariance}
\end{align}
Note that the estimated model $p_{\hat{\vbeta}}$ is possibly unnormalized, if the normalization constant is estimated as well and therefore not exact. Hence, $\mathcal{D}_{\mathrm{KL}}$ here refers to the generalized Kullback-Leibler divergence which is a Bregman divergence and therefore a natural choice for unnormalized models~\citep{matsuda2021unnormalizedmetrics}.

\subsection{Minimizing the Estimation Error}
\label{ssec:parametric_estimation_error_optimization}

The question of statistical efficiency of NCE therefore becomes to optimize Eq.~\ref{eq:gnce_asymp_var}, or its non-parametric counterpart based on Eq.~\ref{eq:avgklerror}, with respect to the four hyperparameters of the classification task: $(\pn, \nu, \phi, T)$. Next, we review what is known of the effects of each of them in detail; the results are mostly about the parametric case in Eq.~\ref{eq:gnce_asymp_var}.

\paragraph{Computation budget, $T$}
First, the effect of the sample budget $T$ on the NCE estimator is clear: following~\eqref{eq:gnce_asymp_var} it scales as $\mathrm{MSE}_{\mathrm{NCE}} \propto \frac{1}{T}$. 
Consequently and remarkably, the optimal values of the other hyperparameters actually do not depend on the budget $T$.

\paragraph{Classification loss, $\phi$}
The effect of the choice of the classification loss ---identified by the function $\phi$ --- has also been solved. Note that it intervenes in the MSE~\eqref{eq:gnce_asymp_var} exclusively through the reweighing function $w(\vx)$.
\citet[Theorem 3]{uehara2018nce} prove that the original logistic loss proposed by~\citet{gutmann2012nce} is in fact optimal, in that it minimizes the estimation error for any fixed value of the other hyperparameters. 
Moreover, this result is consistent with previous work~\citep{pihlaja2010nce} which numerically explored which values of $\phi$ led to the lowest MSE and found a numerical solution that was close to the logistic loss. 
Furthermore, the logistic loss is also numerically stable~\citep{mnih2012condnce} as the density ratio appears inside a sigmoidal function that is bounded and avoids numerical overflow. 
We can accordingly specify $\phi(x) = x\log x - (1+x)\log(1+x)$ in the estimation error \eqref{eq:gnce_asymp_var}. This simplification recovers the estimation error known from~\citet[theorem 3]{gutmann2012nce}, 
\begin{align}
    \mathrm{MSE}_{\mathrm{NCE}}(T, \nu, p_n) 
    &=
    \frac{\nu + 1}{T} \mathrm{tr}
    (
    \mI_w^{-1} - \frac{\nu + 1}{\nu} (\mI_w^{-1} \vm_w \vm_w^\top \mI_w^{-1})
    )
    \label{eq:asympmsence}
\end{align}
where $\vm_w$ and $\mI_w$ are a generalized score mean and covariance, reweighted by misclassified data points $w(\vx) = v(\vx) = P(Y=0 | \vx)$. We will use this formula for the estimation error from now on. It remains to optimize the noise-data ratio $\nu$ and noise distribution $\pn$. 

\paragraph{Classification imbalance, $\nu$}
Regarding the noise-data ratio $\nu$, \citet{gutmann2012nce} and \citet{pihlaja2010nce} report that NCE reaches the Cramer-Rao bound when both $\nu$ and $T$ tend to infinity. However, this is of limited practical use due to finite computational resources $T$. For finite sample size $T$, \citet{pihlaja2010nce} offer numerical results touching on this matter, but in contrast to our framework, they use an algorithm which implicitly assumes a noise proportion to be 50\%.
We present some results on the optimal noise-data ratio in Section~\ref{sec:theoretical_results_general}.

\paragraph{Classification contrast (noise distribution), $p_n$}
Despite some early results, choosing the best noise distribution $\pn$ to reduce the variance of the NCE estimator remains largely unexplored. Like the classification loss, the noise distribution intervenes in the MSE~\eqref{eq:gnce_asymp_var} exclusively through the reweighting functions $w(\vx)$ and $v(\vx)$. It is known that a mismatch (in the tails) of the noise and data distributions can cause the MSE to scale exponentially in the dimension~\citep{lee2022ncegaussiannoise}. It is also known that setting the noise distribution equal to the data's (at least in theory, given the data distribution is unknown) leads to an MSE $(1 + \frac{1}{\nu})$ times higher than the Cramer-Rao bound for normalized models~\citep{pihlaja2010nce,gutmann2012nce}. 
However, it is not known whether choosing $\pn = \pd$ is optimal in terms of MSE in the first place. 
In this paper, our main focus is finding the optimal noise distribution.

\section{Theoretical Results for Computing the Normalization Constant}
\label{sec:theoretical_results_partition_model}

We start by the special case where only the normalization constant $Z$ is ``estimated" (\eqref{eq:unnormalized_partition_model}). 
More general results for estimating any parameters of the data model will be presented in the next section. 

\subsection{Importance Sampling estimators of the Normalization Constant}
\label{ssec:importance_sampling_connection}

We show next that a generalization of NCE recovers well-known Importance Sampling estimators, in the case of computing the normalization constant $Z$ only. 
Importance Sampling is perhaps the most well-known method for computing $Z$~\citep{kahn1949importancesampling,owenmontecarlobook}. It consists in rewriting an integral as an expectation in order to approximate it using a sample from an auxiliary, ``noise" distribution $\pn$ as
\begin{align}
    Z
    &= 
    \int f(\vx) d\vx 
    =
    \E_{\vx \sim \pn}
    \big[ \frac{f(\vx)}{\pn(\vx)} \big] 
    \label{eq:importance_sampling}
\end{align}
This has been generalized by~\citet{meng1996importancesamplingext} in the form
\begin{align}
    Z
    &= 
    \frac{\E_{\vx \sim \pn} [f(\vx) \alpha(\vx)]}
    {\E_{\vx \sim \pd} [\pn(\vx) \alpha(\vx)]}
    \label{eq:importance_sampling_generalized}
     \enspace ,
\end{align}
using both noise \textit{and} data samples, reweighted by a ``hyperparameter'' $\alpha(\vx)$. Choosing $\alpha = \pn^{-1}$ recovers the original Importance Sampling estimator (\eqref{eq:importance_sampling}). 
Other values of $\alpha$ yield other estimators in this extended family, which has motivated important algorithmic developments for estimating $Z$~\citep{gelman1998importancesamplingext}.

Now, we can re-arrange~\eqref{eq:importance_sampling_generalized} 
as
\begin{align}
    0
    &=
    \frac{1}{Z} 
    \E_{\vx \sim \pn} 
    [ f(\vx) \alpha(\vx) ]
      -\E_{\vx \sim \pd} 
    [ \pn(\vx) \alpha(\vx) ]
    \label{eq:is_implicit_formula}
\end{align}
to show it perfectly coincides with \eqref{eq:nce_implicit_alpha} which defines the NCE extension of~\citet[Definition 3]{uehara2018nce}, when the task is balanced $\nu=1$ and the data model is parameterized by the normalization constant alone $ p(\vx; Z) = f(\vx)/Z$. 
We conclude that when the normalization constant $Z$ alone is estimated, the NCE extension of~\citet[Definition 3]{uehara2018nce}, where $\alpha$ is not constrained, actually performs the Importance Sampling extension of~\citet{meng1996importancesamplingext}. 
Thus NCE can be understood as a framework that \textit{extends} the Importance Sampling family of estimators to Energy-Based Models, where both the energy parameters and the normalization constant can be estimated by solving a (self-supervised) classification task. \\

\subsection{Generalized 
NCE estimators of the normalization constant}
\label{ssec:theoretical_results_importance_sampling}

Having shown a correspondence between a family of NCE and Importance Sampling estimators for the normalization constant $Z$, we now turn to individual estimators in the NCE family and show in Table~\ref{table:gnce_normalization} that it is sometimes possible to derive their explicit formulae. 
Choosing the reweighting $\alpha(\vx)$ following~\eqref{eq:nce_implicit_alpha} defines a classification loss.
Minimizing the forward and reverse $\mathrm{KL}$ classification losses recovers the Importance Sampling (IS) and Reverse Importance Sampling (RevIS)~\footnote{this has various names: reverse~\citep{brekelmans2022ebmbounds} or reciprocal~\citep{gelman1998importancesamplingext} importance sampling, or harmonic mean estimator~\citep{newton1994revis}.} estimators of $Z$ ~\citep{newton1994revis}. Minimizing the squared-Hellinger classification loss recovers the product of these estimators (IS-RevIS). These results are obtained by using $p_m(\vx) = \frac{f(\vx)}{Z}$ in Table~\ref{table:classification_loss_generic}, and setting the gradient of the classification loss $\mathcal{L}(Z)$ to zero.
\begin{table*}[tb]
  \caption{
  Binary Classification losses for different choices of convex function $\phi$ recover a family of Importance Sampling estimators for the normalization constant $Z$.
  }
  \centering
  \begin{tabular}{llll}
    \toprule
    Loss & Loss formula (from Table~\ref{table:classification_loss_generic}) & Resulting estimator & Estimator
    \\
    Name & $\mathcal{L}(Z; \phi)$ & $Z = \argmin \mathcal{L}(Z)$ & Name
    \\
    \midrule
    revKL & $-Z \E_{d} \frac{\pn}{f} + \log Z$ & $\left( \E_d \frac{\pn}{f} \right)^{-1}$ & RevIS
    \\
    \midrule
    JS & $\E_{d} \log \sigmoid \left( \log \frac{f}{\nu Z \pn} \right) + \E_{n} \log \sigmoid \left( \log \frac{\nu Z \pn}{f} \right)$ & implicit & NCE
    \\
    \midrule
    KL & $\log Z^{-1} -Z^{-1} \E_{n} \frac{f}{\pn}$ & $\E_n \frac{f}{\pn}$ & IS
    \\
    \midrule
    $H^2$ & $Z^{\frac{1}{2}} \E_d \sqrt{\frac{\pn}{f}} + Z^{\frac{-1}{2}} \E_n \sqrt{\frac{f}{\pn}}$ & $\E_n \sqrt{\frac{f}{\pn}} \left(\E_d \sqrt{\frac{\pn}{f}}\right)^{-1}$ 
    & IS-RevIS
    \\
    \bottomrule
  \end{tabular}
  \label{table:gnce_normalization}
\end{table*}    
Note that the Importance Sampling estimator only uses the noise samples, while the Reverse Importance Sampling estimator only uses the data samples. In contrast, the original NCE estimator which was shown to be optimal~\citep{meng1996importancesamplingext}, employs data samples and noise samples alike. Unlike its counterparts in Table~\ref{table:gnce_normalization}, the original NCE estimator is implicit: it cannot be directly computed using an analytical formula. Instead, it is obtained by numerically minimizing the JS loss for classifying between data and noise samples~\citep{geyer1994ncepartition,gutmann2012nce}.

We next shed some light on the NCE estimator by considering two extreme cases for unbalanced classification: when the budget of samples is limited to one source exclusively, either noise ($\nu \rightarrow \infty$) or data ($\nu \rightarrow 0$). In these limits, we can show (see Appendix~\ref{app:ssec:limits_motivation}) that the JS classification loss becomes the forward and reverse KL classification losses, respectively. These two classification losses correspond to the Importance Sampling and Reverse Importance Sampling estimators which, unlike the original NCE, conveniently possess explicit formulae in table~\ref{table:gnce_normalization}. In other words, we can write
\begin{align}
    \phi_{\mathrm{JS}}(r) 
    &=
    \phi_{\mathrm{KL}}(r)
    + \mathcal{O}_{\nu \rightarrow \infty}(\nu^{-2})
    \\
    \phi_{\mathrm{JS}}(r) 
    &=
    \phi_{\mathrm{revKL}}(r)
    +
    \mathcal{O}_{\nu \rightarrow 0}(\nu)
\end{align}
with the notations from table~\ref{table:classification_loss_generic} and $r = \frac{\pd}{\nu \pn}$. The interpolation between these two limits is performed by the scalar hyperparameter $\nu = \frac{T_n}{T_d}$, which controls the ratio between noise and data observations for a fixed budget $T = T_d + T_n$.  Our observation can be summarized by the following: \textit{NCE interpolates between Importance Sampling in the all-noise limit, and Reverse Importance Sampling in the all-data limit}, when estimating the normalization constant.
This is the cornerstone of our approach in Section~\ref{sec:theoretical_results_general}, where we study the original NCE in these two limits for general parametric models.

\subsection{Optimal noise for estimating the normalization constant}
\label{sec:partition_theoretical_results_parametric}

We next compute the estimation errors given by \eqref{eq:gnce_asymp_var}  for each of the estimators given above. Remarkably, we can express them in terms of well-known divergences 
between the noise and data distributions. The formulae are given in Table~\ref{table:normalization_estimation_error} and proven in Appendix~\ref{app:ssec:estim_error_partition}. 
From the fact that the errors only depend on such divergences, it immediately follows that \textit{the optimal noise distribution is equal to the data distribution} for each of the estimators (\textit{e.g.} IS, NCE, RevIS, IS-RevIS) that we considered. While this reveals the gold standard we should aim for, in practice, the noise distribution cannot be chosen to equal the data distribution as the latter is unknown. The results here are therefore useful for understanding how the difference of data and noise translates into a higher estimation error.  What is more, \textit{the optimal noise proportion is explicitly obtained} for each of these estimators and can be used in practice. Next we consider each case in Table~\ref{table:normalization_estimation_error} separately.

\begin{table*}[!h]
  \caption{
  Estimation errors and optimal noise for different binary classification losses for the estimation of normalization constant $Z$.
  }
  \centering
  \begin{tabular}{llllll}
    \toprule
    Loss & Estimator & Estimator & Estimation Error & \multicolumn{2}{l}{Optimal noise}
    \\
    Name & Name & $Z = \argmin \mathcal{L}(Z)$ & $\mathrm{MSE}(\pn, \nu)$ & $\pn^*$ & $\nu^*$
    \\
    \midrule
    revKL & RevIS & $\left( \E_d \frac{\pn}{f} \right)^{-1}$  & 
    $\frac{1 + \nu}{T} Z^{*^2} \mathcal{D}_{\chi^2}(\pn, \pd)$
    & $\pd$ & $0$
    \\
    \midrule
    JS & NCE & implicit & 
    $\frac{(1 + \nu)^2}{\nu T} Z^{*^2}
    \frac{\mathcal{D}_{\mathrm{HM}}(\pd, \pn)}{1 - \mathcal{D}_{\mathrm{HM}}(\pd, \pn)}$
    & $\pd$ & $1$
    \\
    \midrule
    KL & IS & $\E_n \frac{f}{\pn}$ & 
    $\frac{1 + \nu}{\nu T} Z^{*^2} \mathcal{D}_{\chi^2}(\pd, \pn)$
    & $\pd$ & $\infty$
    \\
    \midrule
    $H^2$ & IS-RevIS & $\E_n \sqrt{\frac{f}{\pn}} \left(\E_d \sqrt{\frac{\pn}{f}}\right)^{-1}$ &
    $\frac{(1 + \nu)^2}{\nu T} Z^{*^2} \frac{\mathcal{D}_{H}^2(\pd, \pn)}{1 - \mathcal{D}_{H}^2(\pd, \pn)}$
    & $\pd$ & $1$
    \\
    \bottomrule
  \end{tabular}
\label{table:normalization_estimation_error}
\end{table*}    

\paragraph{Importance Sampling}
For the Importance Sampling (IS) estimator, obtained by minimizing the KL classification loss, the estimation error deteriorates as the noise distribution grows farther from the data distribution as measured by the chi-squared divergence. This divergence is infinite when the noise has lighter tails than the data.
In practice, for example, image data is heavy-tailed (super-gaussian) so that a Gaussian noise would actually lead to an infinite estimation error. 
In fact, the Importance Sampling estimator is notoriously sensitive to the noise tail and a well-known practical recommendation is that the preferred noise profile is heavy-tailed (\textit{e.g.} Student T rather than a Gaussian).  
The optimal noise proportion for the IS estimator is $100\%$, perhaps unsurprisingly given that the KL loss it minimizes effectively uses noise points only.

\paragraph{Reverse Importance Sampling}
For the Reverse Importance Sampling (RevIS) estimator, obtained by minimizing the reverse KL classification loss, the estimation error grows with the noise-data ``chasm" measured by the reverse chi-squared divergence. This divergence is infinite when the noise is heavier-tailed than the data. 
This leads to the opposite recommendation: the preferred noise profile is light-tailed for the Reverse Importance Sampling estimator~\citep{neal2008revIS}. 
The optimal noise proportion for the revIS estimator is $0\%$, which is intuitive given that the reverse KL loss it minimizes effectively uses data points only.

\paragraph{Product of Importance Sampling and Reverse Importance Sampling}  
The product of the Importance Sampling and Reverse Importance Sampling estimators (IS-RevIS) is obtained by minimizing the Squared Hellinger classification loss. Its estimation error also grows with the noise-data ``chasm" measured by (a function of) the Squared Hellinger distance. 
It is typically finite under a mismatch in the data and noise tails. This suggests it is more robust regarding the choice of the noise distribution. In practice, however, the classification loss used to obtain this estimator is prone to numerical overflow~\citep{liu2021nceoptim}. This is easily seen in table~\ref{table:classification_loss_generic}: for points $\vx$ where the noise and data model distributions share little mass, their ratio $\frac{\pn(\vx)}{p_m(\vx)}$ tends to zero or infinity and so does the (Squared Hellinger) classification loss. 
The optimal noise proportion for the IS-RevIS estimator is $50\%$, indicating that the budget is best divided equally between data and noise points in the classification task.

\paragraph{NCE} The (original) NCE estimator is obtained by minimizing the JS classification loss, which interpolates between the KL and reverse KL losses. Its estimation error grows with the noise-data ``chasm" measured by (a function of) the harmonic divergence, rather than the KL divergence as conjectured in~\citep{rhodes2020bridgedre}. 
The estimation error of NCE is also typically finite under a mismatch in the data and noise tails.
Furthermore, its classification loss is numerically stable~\citep{mnih2012condnce}. This is easily seen in table~\ref{table:classification_loss_generic}: the ratio $\frac{\pn(\vx)}{p_m(\vx)}$ is transformed by a sigmoidal function that bounds its range to $[0, 1]$.  
In this sense, the original NCE estimator is robust to, and even optimal~\citep[Theorem 3]{uehara2018nce} regarding the choice of the noise distribution.
The optimal noise proportion for the NCE estimator is $50\%$, indicating that the budget is best divided equally between data and noise points in the classification task.

This begs the question: is the optimal noise distribution still equal to the data distribution when we estimate \textit{both} the energy parameters and the normalization? is the optimal noise proportion still 50\% for the original NCE? and what does the estimation error look like?

\section{Theoretical Results for any Parametric Model}
\label{sec:theoretical_results_general}

We now proceed to our main results for estimating a general parametric model of the data, whether normalized or unnormalized. 
We aim to directly minimize the estimation error (MSE) of the NCE estimator with respect to the noise distribution. Though the estimation error is analytically known in \eqref{eq:gnce_asymp_var},
its minimization with respect to the noise distribution $p_n$ is a difficult task.
Even in the simple case where the data follows a one-dimensional Gaussian distribution parameterized by variance, the resulting expression for the estimation error is intractable as explained in Appendix~\ref{ssec:intractability}. 

Different approaches can be taken to face this difficulty. 
\citet{lee2022ncegaussiannoise} resorted to bounding the estimation error's dependency on the dimension in a simple setting when the noise and data have different tails. The approach we take in this section assumes nothing on the data and noise distributions: instead, we consider the all noise and all data limits introduced in Section~\ref{ssec:theoretical_results_importance_sampling}, which will make the minimization of the estimation error tractable.
In Sections~\ref{sec:theoretical_results_parametric} and ~\ref{sec:theoretical_results_nonparametric}, we theoretically derive the noise distributions which minimize the parametric and non-parametric estimation errors in these limits. 
We obtain that an optimal noise distribution is a reweighted version of the data distribution --- not the data distribution itself. The reweighting emphasizes data points $\vx$ where the classifier is most sensitive to its estimated parameters $\hat{\vbeta}$. 

\subsection{Minimizing the Parametric Error}
\label{sec:theoretical_results_parametric}

In the first limit case of all noise observations, we are able to solve for the noise minimizing the estimation error (MSE). For notational simplicity, define $f \bullet g := fg / \int fg$ as a normalized product operator. We recall that $\mI$ is the covariance of the (generalized) Fisher score vector, defined in~\eqref{eq:unweighted_covariance}. Our main result is the following theorem:
\begin{theorem}[Optimal noise distribution, all noise limit] 
    \label{th:allnoisebestmse}
    In the limit of all noise observations, $\nu \rightarrow \infty$, the parametric estimation error depends on the noise via a forward $\chi^2$ divergence 
    \begin{align}
        \mathcal{D}_{\chi^2}( p_d \bullet w_1, p_n)
    \end{align}
    where the data reweighting function
    \begin{align}
        w_1(\vx) 
        = 
        \| \fisherscore[\vx] \|_{\mI^{-2}}
        \label{eq:allnoisemsereweighting}
    \end{align}
    emphasizes data points whose classification is most sensitive to the learnt classifier $F(\vx; \vbeta^*)$.
    The noise distribution minimizing the parametric estimator error is thus a reweighted version of the data distribution
    \begin{align}
        p_n^{\mathrm{opt}}(\vx) 
        & = 
        (p_d \bullet w_1)(\vx)
        \enspace .
        \label{eq:allnoisebestmse} 
    \end{align}
\end{theorem}
Note that $\nabla_{\vbeta} F(\vx; \vbeta^*)$ is equal to the Fisher score generalized by including the normalization constant (as shown in Eq.~\ref{eq:Flogp} for the normalized case). Also note that the notation $\langle \vx, \vy \rangle_{\mA} := \langle \vx, \mA \vy \rangle$ refers to the inner product with metric $\mA$. The induced norm is $\| \vx \|_{\mA} := \| \mA^{\frac{1}{2}} \vx \|$. This optimal noise distribution matches that of~\citet{pihlaja2010nce} obtained for the Importance Sampling estimator. This is not surprising, given that NCE is equivalent to Importance Sampling in the all noise limit as explained in Section~\ref{ssec:theoretical_results_importance_sampling}.

The proof of Theorem~\ref{th:allnoisebestmse} is available in Appendix~\ref{app:ssec:proof_mse_allnoise_parametric}, where we also show it holds for another, alternative assumption: when the noise distribution is a (infinitesimal) perturbation of the data distribution $\frac{p_d}{p_n}(\vx) = 1 + \eps(\vx)$. 

In the second limit of all data observations, we can more informally characterize the noise distribution which minimizes the parametric estimation error.
\begin{conjecture}[Optimal noise distribution, all data limit]
    \label{conj:one}
    Consider a scalar parameter $\beta \in \mathbb{R}$. 
    In the limit of all data observations, $\nu \rightarrow 0$, an optimal noise distribution concentrates its mass in regions where the classifier's parameter-sensitivity $\nabla_{\beta} F(\vx; \beta^*)$ --- or Fisher score ---   is constant. Such noise distributions affect the parametric error via a reverse $\chi^2$ divergence 
    \begin{align}
        \mathcal{D}_{\chi^2}(p_n,  p_d \bullet w_2)
        \label{eq:mse_alldata_param}
    \end{align}
    where the data reweighting function is given by
    \begin{align}
        w_2(\vx) 
        &= 
        \big(
        \nabla_{\vbeta}
        F(\vx; \beta^*)
        \big)^2 
        \enspace .
        \label{eq:alldatabestmse}
    \end{align}
    The noise distribution minimizing the parametric error is therefore concentrated at the modes of a reweighted version of the data distribution $p_d \bullet w_2$
    that share the same classifier parameter-sensitivity $\nabla_{\beta}F(\vx; \beta^*)$. There can be as many equivalent optimal noise distributions there are modes corresponding to different parameter-sensitivities of the classifier. 
    \label{th:alldatabestmse}
\end{conjecture}
The optimal noise distribution is typically degenerate,
consisting of a Dirac delta or perhaps more than one of them. 
The ``proof" of this conjecture is not quite rigorous precisely due to such degeneracy, which is why we label this as conjecture only. 

How do these general results compare to those we obtained in  Section~\ref{sec:partition_theoretical_results_parametric} for estimating the normalization constant only? Similar to the estimation of $Z$, the deviation from optimality is characterized using divergences: forward and reverse chi-squared divergences in the present all noise and all data limits. However, here the divergences involve a reweighted version of the data distribution, not the original data distribution as when estimating $Z$ only. Therefore, the optimal noise distribution is not the data distribution itself. We note that the reweighting involves the score of the parameter, and that this score is constant when estimating the 
log-normalization constant alone (see Section \ref{ssec:parametric_models}), which explains the difference in the results. In the general case, the score emphasizes data points $\vx$ where the classifier is most sensitive to its estimated parameters $\hat{\vbeta}$.

In practice, the results above mean that the optimal noise distribution is even harder to sample from than the data distribution: we cannot just re-use data observations. Its density is also harder even to evaluate, as it requires knowing the estimated data model.
This begs the question: to what extent is targeting the optimal noise worth the trouble? While this is a rather empirical question, we can approach an answer by the following theorem: 
\begin{theorem}[Deviation from optimality, all noise limit] 
    \label{th:mse_gap_allnoise}
    In the all noise limit of Theorem~\ref{th:allnoisebestmse}, the gap between the MSE for the typical case $p_n = p_d$ and the optimal case $p_n = p_n^{\mathrm{opt}}$ is given by
    \begin{align}
        \mathrm{MSE}(\pd) - \mathrm{MSE}(p_n^{\mathrm{opt}})
        &= 
        \frac{1}{T} 
        \Var_{x \sim \pd}
        w_1(\vx)
        \label{eq:allnoise_adversarial_gap}
        \enspace .
    \end{align}
    where the reweighting function $w_1 (x)$ is defined in~\eqref{eq:allnoisemsereweighting}.
    
    Furthermore, for normalized models, we can compute the gap to efficiency in the all noise limit, i.e.\ between $p_n = p_n^{\mathrm{opt}}$ and the Cramer-Rao lower bound 
    \begin{align}
        \mathrm{MSE}(p_n^{\mathrm{opt}})
        - 
        \mathrm{MSE}_{\mathrm{Cramer-Rao}}
        = 
        \frac{1}{T} \mathbb{E}_{x \sim p_d}(
        \| 
        \fisherscore[x]
        \|_{\mI^{-2}}
        )^2
        \enspace .
    \end{align}
\end{theorem}
The gap in \eqref{eq:allnoise_adversarial_gap} seems to be positive for any reasonable distribution, which implies that the optimal noise cannot be the data distribution $\pd$.  
However, we should note that the gain in efficiency provided by the optimal noise distribution can be modest compared to the special case $\pn = \pd$, since we know from~\citep{pihlaja2010nce,gutmann2012nce} that the latter setting achieves $(1 + \frac{1}{\nu})$ times the Cramer-Rao bound. 

Finally, we consider optimization of the noise proportion. It is often heuristically  assumed that having 50\% noise, i.e.\ $\nu=1$ is optimal. We have shown that this is true when estimating the normalization only in Section~\ref{sec:partition_theoretical_results_parametric}: does this hold when estimating any parameterization of the model distribution? In the special case when $\pn = \pd$, we can actually show in Appendix~\ref{app:sec:gnce_optimal_noise_proportion} that the optimal noise proportion is $50\%$\footnote{This is obtained for a fixed computational budget $T$, which is our approach in this paper. \citet{gutmann2012nce} consider the case where this constraint on the budget is relaxed, and show that the optimal noise proportion in their $\nu \rightarrow \infty$ as in Corollary 7 and Figure 4.d.}.
The converse of this result is not true: a noise proportion of $50\%$ does \textit{not}
ensure that the noise distribution equals the data's, as we will numerically illustrate in Section~\ref{sec:numerical_results}.

\subsection{Minimizing the Non-parametric Error}
\label{sec:theoretical_results_nonparametric}

We next adapt our results to the case of the non-parametric estimation error.
In the first limit of all noise observations, we are able to solve for the noise minimizing the non-parametric estimation error:
\begin{theorem}[Optimal noise distribution, all noise limit] 
    \label{th:allnoisebestkl}
    In the limit of all noise observations $\nu \rightarrow \infty$, the non-parametric estimation error depends on the noise via a forward $\chi^2$ divergence 
    \begin{align}
        \mathcal{D}_{\chi^2}( p_d \bullet w_3, p_n)
    \end{align}
    where the data reweighting function
    \begin{align}
        w_3(\vx) 
        = 
        \| \fisherscore[\vx] \|_{\mI^{-1}}     
        \label{eq:allnoiseklreweighting}
    \end{align}
    emphasizes data points whose classification is most sensitive to the learnt classifier $F(\vx; \vbeta^*)$.
    The noise distribution minimizing the parametric estimator error is a reweighted version of the data distribution
    \begin{align}
        p_n^{\mathrm{opt}}(\vx) 
        & = 
        (p_d \bullet w_3)(\vx)
        \enspace .
        \label{eq:allnoisebestkl} 
    \end{align}
\end{theorem}
Again, the proof of Theorem~\ref{th:allnoisebestkl} is available in Appendix~\ref{app:ssec:proof_mse_allnoise_parametric}, where we also show it holds for another, alternative assumption: when the noise distribution is a (infinitesimal) perturbation of the data distribution $\frac{p_d}{p_n}(\vx) = 1 + \eps(\vx)$. 
In the second limit of all data observations, we can again informally characterize the noise distribution which minimizes the non-parametric estimation error.
\begin{conjecture}[Optimal noise distribution, all data limit]
    \label{conj:two}
    The following holds for a scalar parameter $\beta \in \mathbb{R}$.
    In the limit of all data observations $\nu \rightarrow 0$, 
    an optimal noise distribution concentrates its mass in regions where the classifier's parameter-sensitivity  $\nabla_{\beta} F(\vx; \beta^*)$ --- or Fisher score --- is constant. Such noise distributions affect the non-parameteric error via a reverse $\chi^2$ divergence    
    \begin{align}
        \mathcal{D}_{\chi^2}(p_n,  p_d \bullet w_4)
    \end{align}
    where the data reweighting function is
    \begin{align}
        w_4(\vx) 
        &= 
        |\nabla_{\beta}
        F(\vx; \beta^*)|    
        \label{eq:alldatabestkl}
    \end{align}
    for a scalar parameter $\beta$. It emphasizes data points whose classification is most sensitive to the learnt classifier $F(\vx; \beta^*)$.
    Specifically, the noise distribution minimizing the parametric error is concentrated at the modes of a reweighted version of the data distribution $p_d \bullet w_4$ that share the same parameter-sensitivity $\nabla_{\beta}F(\vx; \beta^*)$. There can be as many equivalent optimal noise distributions as there are modes with different classifier parameter-sensitivities. 
\label{th:alldatabestkl}
\end{conjecture}
These optimal noise distributions resemble those from Theorem~\ref{th:allnoisebestmse} and Conjecture~\ref{th:alldatabestmse}: only the exponent on the generalized Fisher Information matrix changes. This changes the \textit{curvature} used to measure the reweighting in the parametric distribution space. This is predictable, as the new cost function $\frac{1}{2T_d} \mathrm{tr}(\mSigma \mI)$ is obtained by scaling with the generalized Fisher Information matrix. More specifically, when the data parameter is scalar, the optimal noises from Theorems~\ref{th:allnoisebestmse} and \ref{th:allnoisebestkl} coincide, as the generalized Fisher Information becomes a multiplicative constant; those from Conjectures~\ref{th:alldatabestmse} and \ref{th:alldatabestkl} do not coincide but are rather similar.

\section{Numerical results: optimal noise distribution and proportion}
\label{sec:numerical_results}

We now turn to experiments to validate the theory above. Specifically, we verify our formulae for the optimal noise distribution of the original NCE, in the all-data~(Eq.~\ref{eq:alldatabestmse}) and all-noise~(Eq.~\ref{eq:allnoisebestmse}) limits, by numerically minimizing the MSE~(Eq.~\ref{eq:asympmsence}). Outside these limits, we show that the optimal noise is an interpolation between both limits and that our formulae decrease the MSE compared with setting the noise distribution equal to the data's. 

\subsection{Numerical Methods} 
The integrals in Eq.~\ref{eq:gnce_asymp_integrals} required for evaluating the asymptotic MSE
can be approximated using numerical integration (quadrature) or Monte-Carlo simulations. While both approaches lead to comparable results, quadrature is significantly faster and more precise for very low-dimensional data. However, using Monte-Carlo leads to an estimate that is fully differentiable with respect to the parameters of $\pn$. 

In our simulations, the noise distribution $\pn$ is either constrained to the same \textit{parametric} family as the data distribution, using a single scalar parameter; or else is a histogram able to approximate \textit{non-parametric} distributions. In the histogram model, each bin is assigned a parameter except for the last bin whose mass is computed based on the normalization constraint.
To tackle the one-dimensional parametric problem (Figs.~\ref{fig:noiseval_vs_dataval},~\ref{fig:msevsnoiseprop}), we simply employed quadrature for evaluating the function to optimize over a dense grid and then selected the minimum. This appeared as the most computationally efficient strategy and allows for visualizing the MSE landscape reported in Appendix~\ref{ssec:intractability}.
In the multi-dimensional non-parametric case (Figs.~\ref{fig:optimalnoisemean},~\ref{fig:optimalnoisevar},~\ref{fig:optimalnoisecorr}), the histogram's weights can be optimized by first-order methods using automatic differentiation applied to Monte Carlo estimates.

In the following experiments, the optimization strategy consists in obtaining the gradients of the Monte-Carlo estimate using PyTorch~\citep{pytorch} and plugging them into a non-linear conjugate gradient scheme implemented in Scipy~\citep{scipy}. We chose the conjugate-gradient algorithm as it is deterministic 
(unlike SGD) and offered fast convergence toward numerically precise minimizers. None of the experiments below required more than 100 iterations of conjugate-gradient. 
Note that for numerical precision, we had to set PyTorch's default to 64-bit floating-point precision.
Our code is available at \href{https://github.com/l-omar-chehab/nce-noise-var}{https://github.com/l-omar-chehab/nce-noise-var}.

\subsection{Toy data}
\label{ssec:toy_data}

The data distributions considered in our experiments are picked among three generative models with a scalar parameter:
\begin{center}
\begin{tabular}{ lll } 
\toprule
Model 
& 
Unnormalized distribution $\tilde{p}(x; \theta)$
& 
Normalization $Z(\theta)$
\\ 
\midrule
(a) Gaussian Mean (1D)
& 
$\exp
\left(
\frac{-1}{2} (x  - \theta)^2
\right)$
& 
$\frac{1}{\sqrt{2\pi}}$
\\ 
\midrule
(b) Gaussian Var (1D)
& 
$\exp
\left(
\frac{-1}{2} x^2 \theta^{-1} 
\right)$
& 
$\frac{1}{\sqrt{2\pi\theta}}$
\\ 
\midrule
(c) Gaussian Correlation (2D)
& 
$\exp
\left(
\frac{-1}{2} \vx^\top 
\begin{bmatrix}
1 & \theta \\
\theta & 1
\end{bmatrix}^{-1}
\vx
\right)$
& 
$\frac{1}{\sqrt{2\pi (1 - \theta^2)}}$
\\ 
\bottomrule
\end{tabular}
\end{center}
In each of these cases, whenever unspecified, the mean is zero and the marginal variances fixed to one. We use here simple data distributions to illustrate the difficulty of finding the optimal distribution. While the Gaussian distribution is very simple, it is in fact ubiquitous in generative models literature.
Yet, to our knowledge, there is little work addressing how to design the optimal noise to infer the parameters of a Gaussian using NCE. 

\subsection{Estimating Normalized Models}
\label{ssec:model_normalized}

In the following, we first present numerical results on estimating the normalized models (a), (b), and (c) defined in Section~\ref{ssec:toy_data}.

\begin{figure}[!ht]
\centering
\includegraphics[width=\columnwidth]{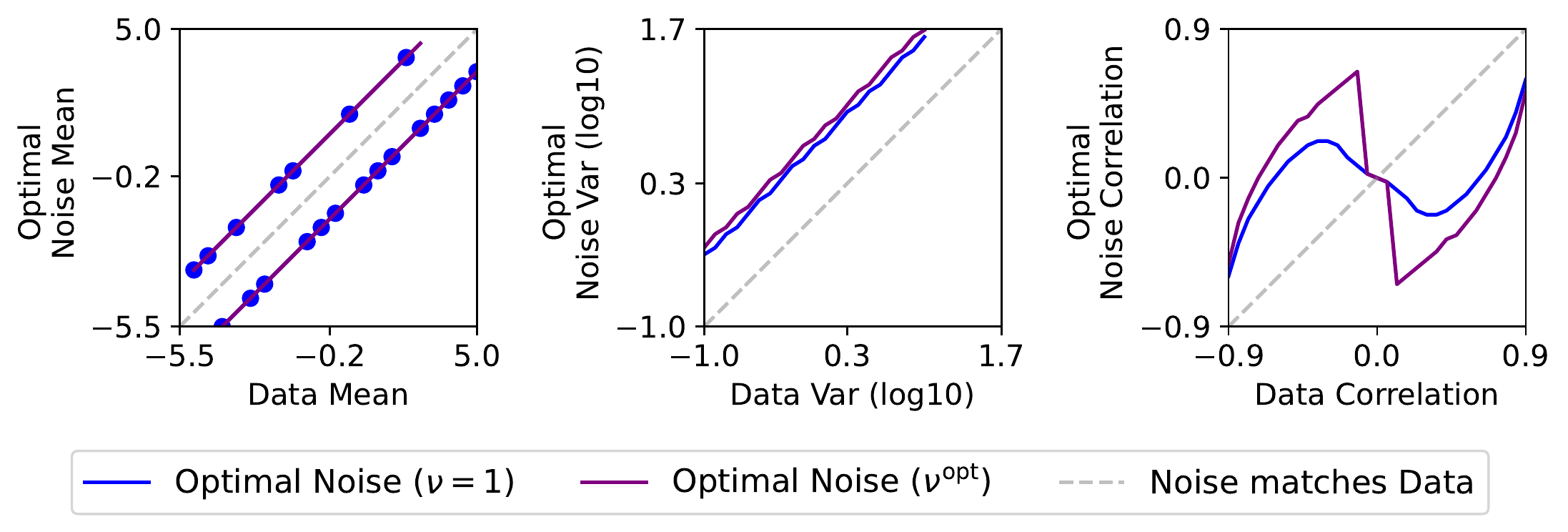}
\caption{Relationship between the (optimal) noise parameter and the data parameter of a normalized model. The optimal noise parameter is obtained when the noise proportion is fixed at 50\% $(\nu = 1)$ or jointly minimized $\nu = \nu^{opt}$. 
(Top left) Optimal variance in model (a) as function of the data mean. Note that the noise parameter has two symmetric local minima; to clarify this, we plot the results of individual runs as blue points, which are joined by two manually drawn lines. (Top right) Optimal variance in model (b) as function of the data variance. (Bottom left) Optimal noise correlation in model (c) as a function of the data correlation.}
\label{fig:noiseval_vs_dataval}
\end{figure}

\paragraph{Parametric Noise Distribution}
We start by assuming the same parametric distribution for the noise as for the data, and consider parameter estimation in normalized models.
Figure~\ref{fig:noiseval_vs_dataval} presents the optimal noise parameter as a function of the data parameter for normalized models. 
For the three models above, we set the noise proportion to 50\% (i.e. $\nu = 1$) or jointly minimize it (i.e. $\nu = \nu^{opt}$) along with the noise parameter. One can observe that the optimal noise parameter systematically differs from the data parameter. They are equal only in the very special case of estimating correlation (case c) for uncorrelated variables.
This means that the optimal noise distribution is not equal to the data distribution even when the noise and the data are restricted to be in the same parametric family of distributions. This is coherent with our result in~\eqref{eq:allnoise_adversarial_gap}.

Looking more closely, one can notice that the relationship between the optimal noise parameter and the data parameter highly depends on the estimation problem. For model (a), the optimal noise mean is (randomly) above or below the data mean, 
while at constant distance. These are two global minima of the MSE landscape shown in Appendix~\ref{ssec:intractability}. For model (b), the optimal noise variance is obtained from the data variance by a scaling of $3.84$. This linear relationship is coherent with the symmetry of the problem with respect to the variance parameter. Interestingly for model (c), the optimal noise parameter exhibits a nonlinear relationship to the data parameter: for a very low positive correlation between variables the noise should be negatively correlated, whereas when data variables are strongly correlated, the noise should also be positively correlated. 

\paragraph{Unconstrained Noise Distribution}
Next we consider the non-parametric optimization of the noise distribution, without constraining it in any way.

Before showing the results, we point out that we can derive an optimal noise distribution for model (a) in closed form. In the all data limit, minimizing the MSE in Eq.~\ref{eq:mse_alldata_param}  yields two candidates for $p_n^{\mathrm{opt}}(x)$ to concentrate its mass on: $\delta_{-\sqrt{2}}$ and $ \delta_{\sqrt{2}}$. Moreover, our Conjecture~\ref{th:alldatabestmse}
predicts how the probability mass should be distributed to the two candidates: because they have different scores, they are two distinct global minima. This is coherent with the two minima observed for the Gaussian mean in Figure~\ref{fig:noiseval_vs_dataval} (top-left). Similarly, when estimating a Gaussian variance, minimizing the MSE in Eq.~\ref{eq:mse_alldata_param} yields candidates $\delta_{-\sqrt{5}}$ and  $\delta_{\sqrt{5}}$ for $p_n^{\mathrm{opt}}(x)$. In this case however, both candidates have the same score.
Our theory in Conjecture~\ref{th:alldatabestmse} does not say anything about how the probability mass should be distributed to these two points: it can be 50-50 or all on just one point. A possible solution is $p_n^{opt}(x) =
\frac{1}{2}(\delta_{-\sqrt{5}} + \delta_{\sqrt{5}})$.

\begin{figure}[!ht]
\centering
\begin{subfigure}[t]{0.48\textwidth}
    \centering
    \includegraphics[width=0.9\linewidth]{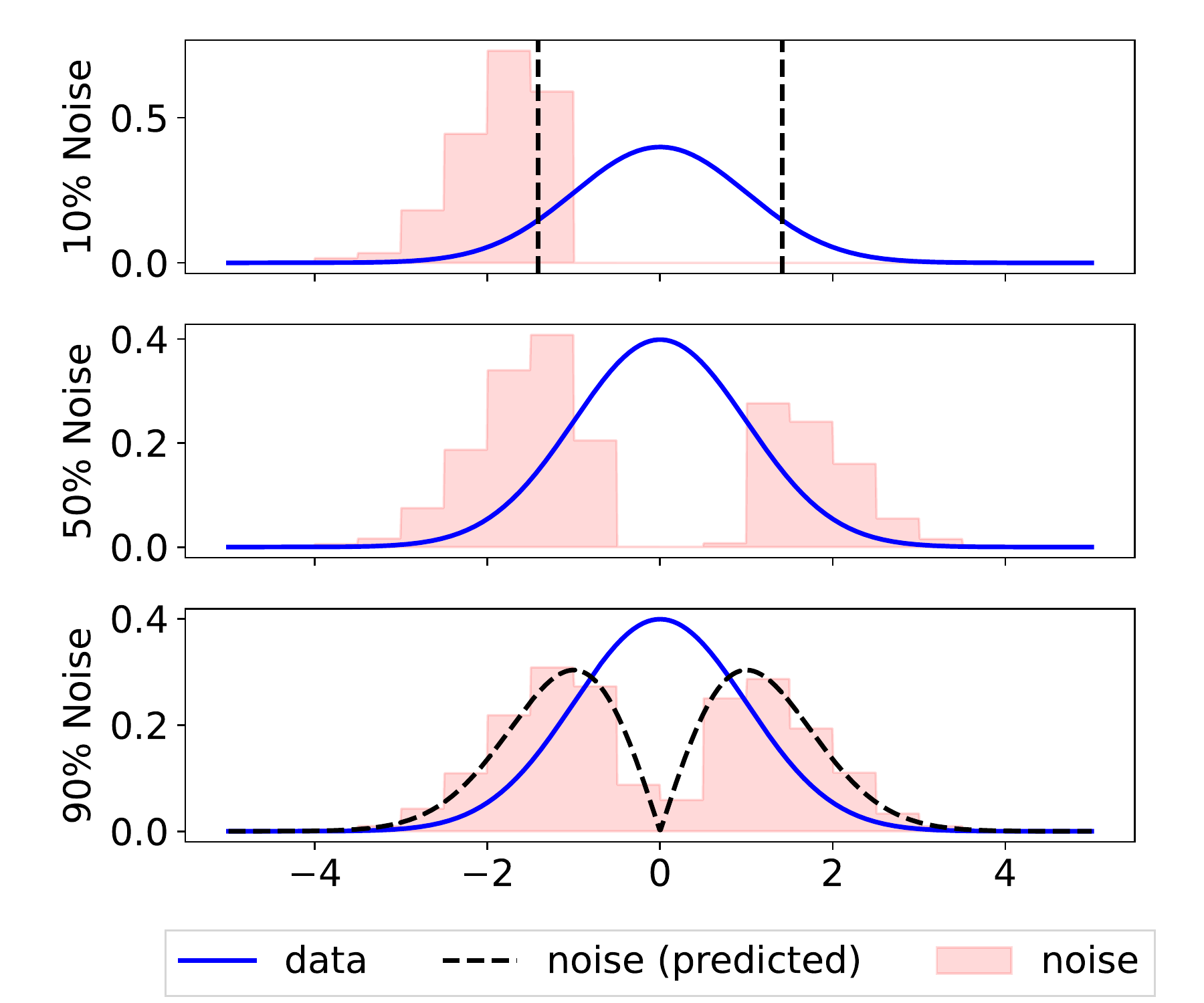} 
    \caption{Model a: Gaussian mean}
    \label{fig:optimalnoisemean}
\end{subfigure}
\hfill
\begin{subfigure}[t]{0.48\textwidth}
    \centering
    \includegraphics[width=0.9\linewidth]{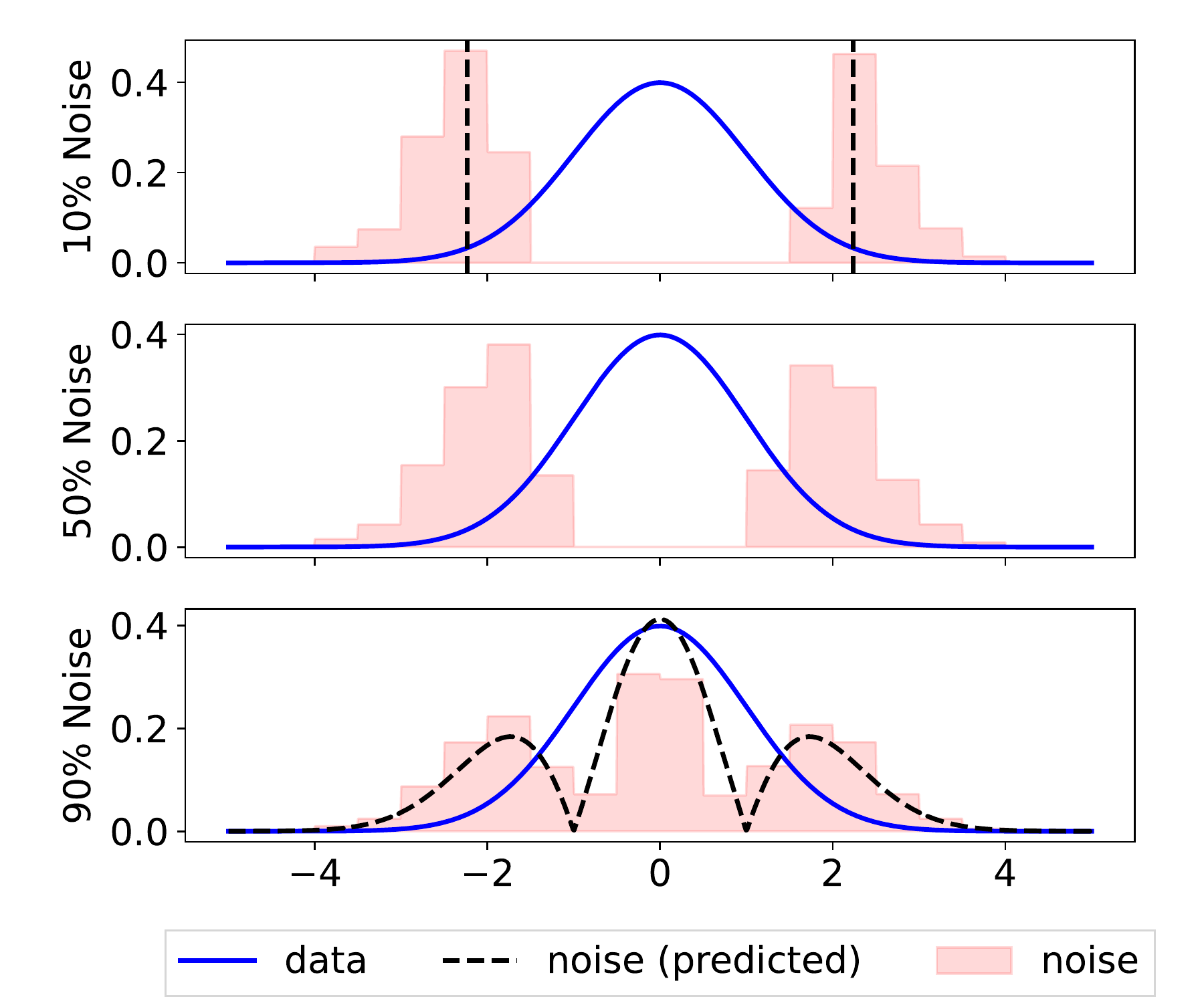} 
    \caption{Model b: Gaussian variance}
    \label{fig:optimalnoisevar}
\end{subfigure}
\caption{Unconstrained (nonparametric histogram-based) optimal noise distributions when the data model is normalized. Each row gives a different $\nu$ or noise proportion. The pink bars give the numerical approximations of the optimal noise distribution. The theoretical approximation of optimal noise is given by the dashed lines: the all-noise limit in the bottom panel, and the all-data limit in the top panel. In the top panel, the optimal noise is given by single points (Dirac masses) which are chosen symmetric for the purposes of illustration, but as explained in the text, they are two global minima in the case of Gaussian mean estimation, whereas when estimating the variance, any distribution of probability on those two points is equally optimal.}
\end{figure}

Figure \ref{fig:optimalnoisemean} shows the optimal histogram-based noise distribution for estimating the mean of a Gaussian, together with our theoretical predictions (Theorem~\ref{th:allnoisebestmse} and Conjecture~\ref{th:alldatabestmse}).
We can see that our theoretical predictions in the all-data and all-noise limits match numerical results. It is apparent in  Figure~\ref{fig:optimalnoisemean} that the optimal noise places its mass where the data distribution is high, and where it varies most when $\theta^*$ changes.
Furthermore, the noise distribution in the all-data limit has higher mass concentration, which also matches our predictions. Interestingly, in a case not covered by our hypotheses, when there are as many noise observations as data observations, i.e. noise proportion of 50\% or $\nu = 1$, the optimal noise in Figure~\ref{fig:optimalnoisevar} (middle) is qualitatively not very different from the limit cases of all data or all noise observations.
It is here important to take into account the indeterminacy of distributing probability mass on the two Diracs, which is coherent with initial experiments in Figure~\ref{fig:noiseval_vs_dataval} as well as the MSE landscape included in Appendix~\ref{ssec:intractability}.
Figure \ref{fig:optimalnoisemean} is a perfect illustration of a complex phenomenon occurring in a setup as simple as Gaussian mean estimation.
Our conjecture in Eq.~\ref{eq:alldatabestmse} predicts the equivalent optimal noises seen in our experiments, in Figure 1 (top-left) and Figure 2.b., where the noise concentrates its mass on either point of the set $\{-\sqrt{2}, \sqrt{2}\}$. Indeed, Eq.~\ref{eq:alldatabestmse} shows that any noise which concentrates its mass on a set of points where the score is constant is (equally) optimal. So despite its approximative quality, Eq.~\ref{eq:alldatabestmse} is able to explain what we observed empirically: in the all-data limit, there can be many equivalent optimal noises.

Figure~\ref{fig:optimalnoisevar}
gives the same results for the estimation of a Gaussian's variance. The conclusions are similar.

\begin{figure}[!t]
\centering
\includegraphics[width=0.55\columnwidth]{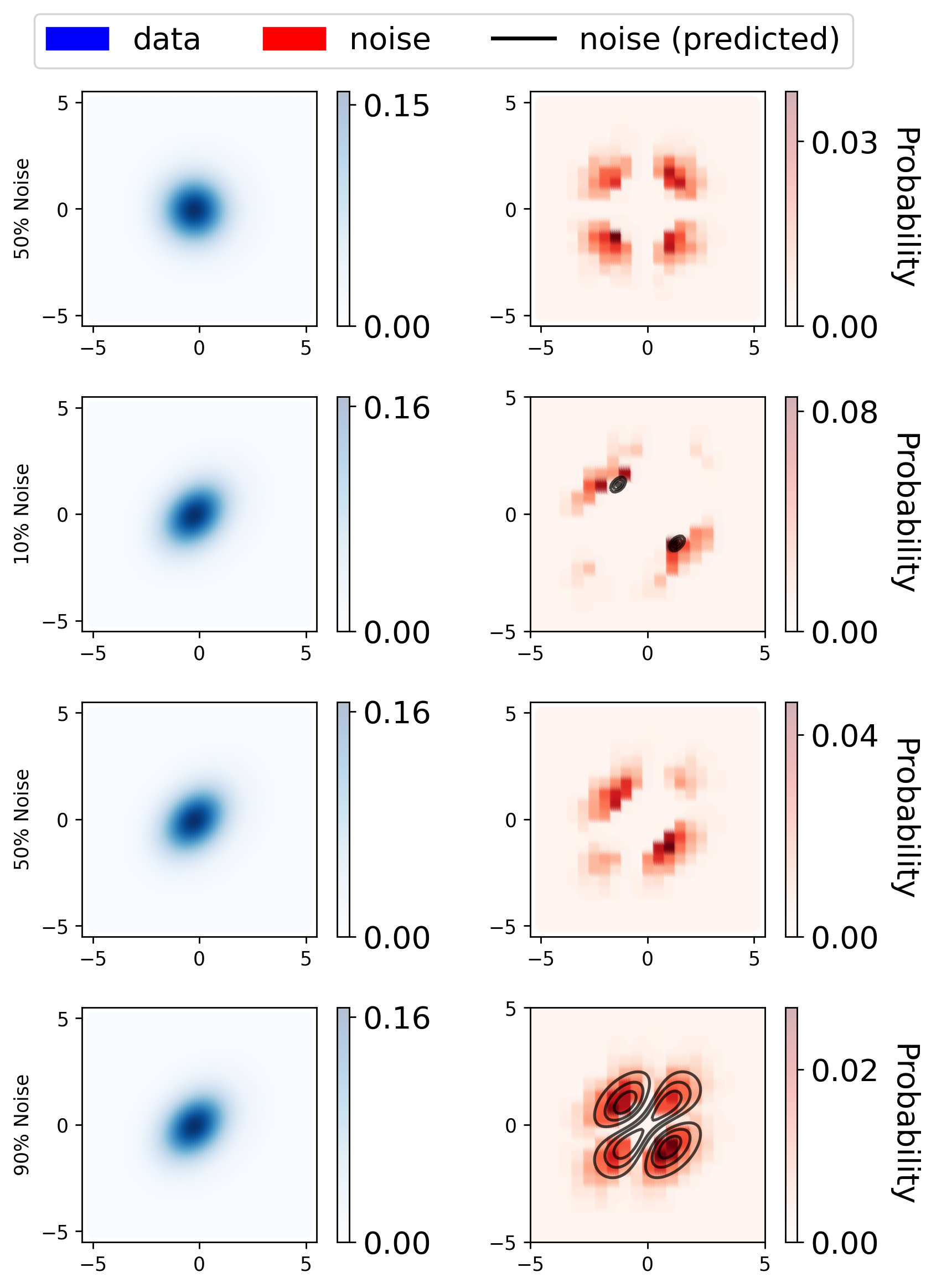}
\caption{Optimal noise for a 2D Gaussian parameterized by correlation (model c). 2D Gaussian with correlation 0 (top row) and 0.3 (next three rows) are considered. Left panel is data density, right panel is the optimal histogram-based noise density.
The theoretical approximation of optimal noise is given by the black level lines: the case of Theorem~\ref{th:allnoisebestmse} in the bottom panel, and the Conjecture~\ref{conj:one} in the second panel. Here, the optimal noise in the latter limit is given by a softmax relaxation with temperature 0.01. It makes the choice of placing its mass symmetrically on the single points (Dirac masses), but as explained in the text, any distribution of probability on those two points could be equally optimal.} 
\label{fig:optimalnoisecorr}
\end{figure} 

Figure~\ref{fig:optimalnoisecorr} shows the numerically estimated optimal noise distribution for model (c) using a Gaussian correlation parameter. Here, the distributions are perhaps even more surprising than in previous figures. This can be partly understood by the extremely nonlinear dependence of the optimal noise parameter from the data parameter shown in Fig.~\ref{fig:noiseval_vs_dataval}.

We conclude that throughout, the optimal noise distributions are highly \textit{non-Gaussian unlike the data distribution}. 

\begin{figure}[!t]
\centering
\includegraphics[width=0.6\columnwidth]{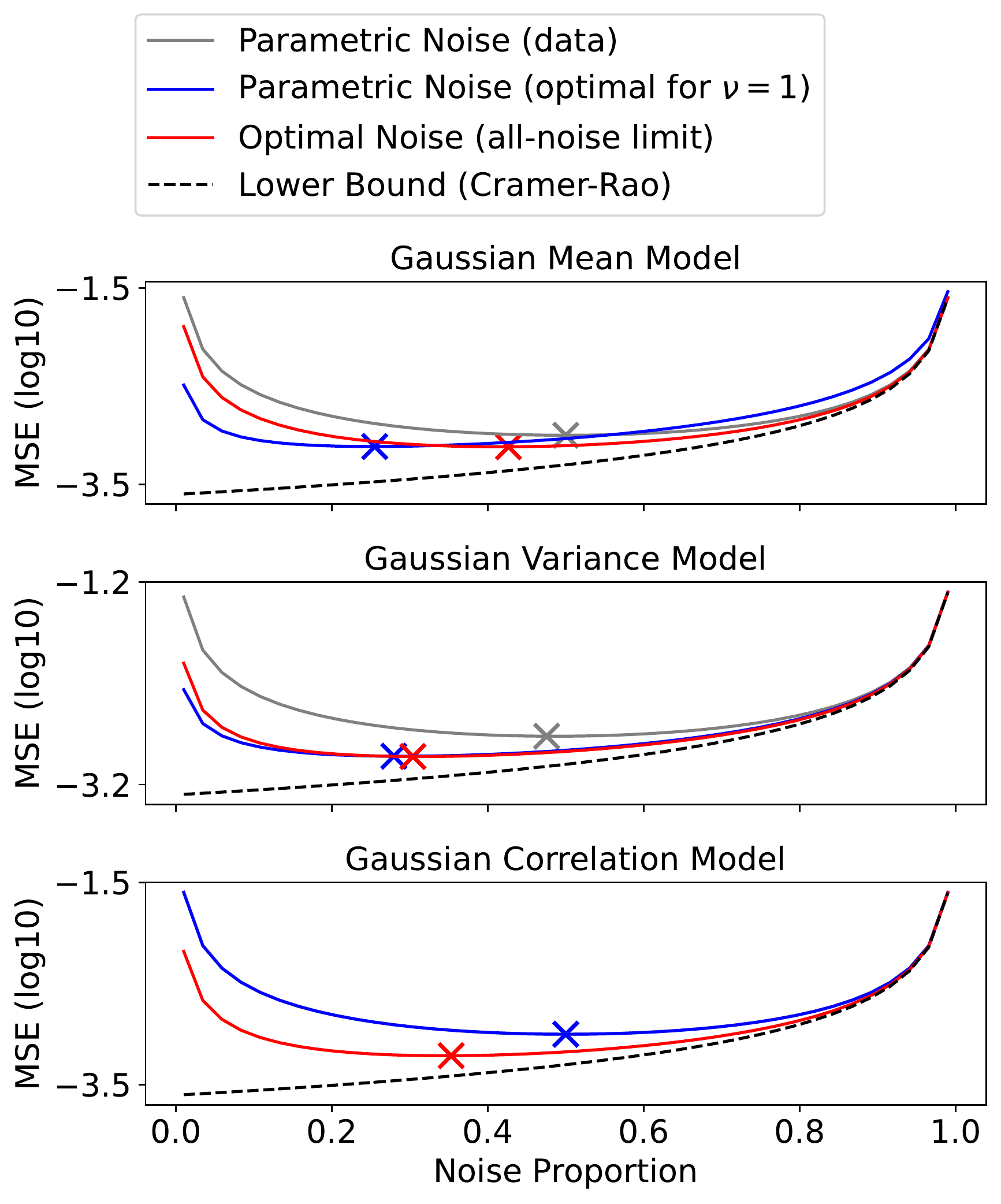}
\caption{Asymptotic MSE vs. noise proportion. Top panel: Asymptotic MSE vs. noise proportion for model (a) with parameter mean; Middle panel: Asymptotic MSE vs. noise proportion for model (b) with parameter variance; Bottom panel: Asymptotic MSE vs. noise proportion for model (c) with parameter correlation \aapo{(the gray curve is not visible because it is coincides with the blue curve)}. \aapo{``Parametric noise (data)" means that the noise distribution equals the data distribution.} The parameter in ``parametric noise" is the optimal parameter for $\nu=1$, i.e. for when half the observations are noise and half are data. The ``optimal noise" is the approximation given by Theorem~\ref{th:allnoisebestmse}.
}
\label{fig:msevsnoiseprop}
\end{figure}

\paragraph{Robustness of our theoretical results}
We next ask: how robust to $\nu$ is the analytical noise we derived in these limiting cases?
Figure~\ref{fig:msevsnoiseprop} shows the Asymptotic MSE achieved by two noise models, across a range of noise proportions. The first noise model is the optimal noise in the parametric family containing the data distribution $p_n = p_{\theta}$, optimized for $\nu=1$, while the second noise model is the optimal analytical noise $p_n^{\mathrm{opt}}$ derived in the all-noise limit~(Eq.\ref{eq:allnoisebestmse}). They are both compared to the Cramer-Rao lower bound. For all models (a) (b) and (c), the optimal analytical noise $p_n^{\mathrm{opt}}$ (red curve) is empirically useful even far away from the all-noise limit, and across the entire range of noise proportions. In fact, $p_n = p_n^{\mathrm{opt}}$ empirically seems a better choice than using the data distribution $p_n = p_d$, and is (quasi) uniformly equal to or better than a parametric noise $p_n = p_\theta$ optimized for $\nu=1$.

\begin{figure}[!ht]
\centering
\includegraphics[width=0.6\columnwidth]{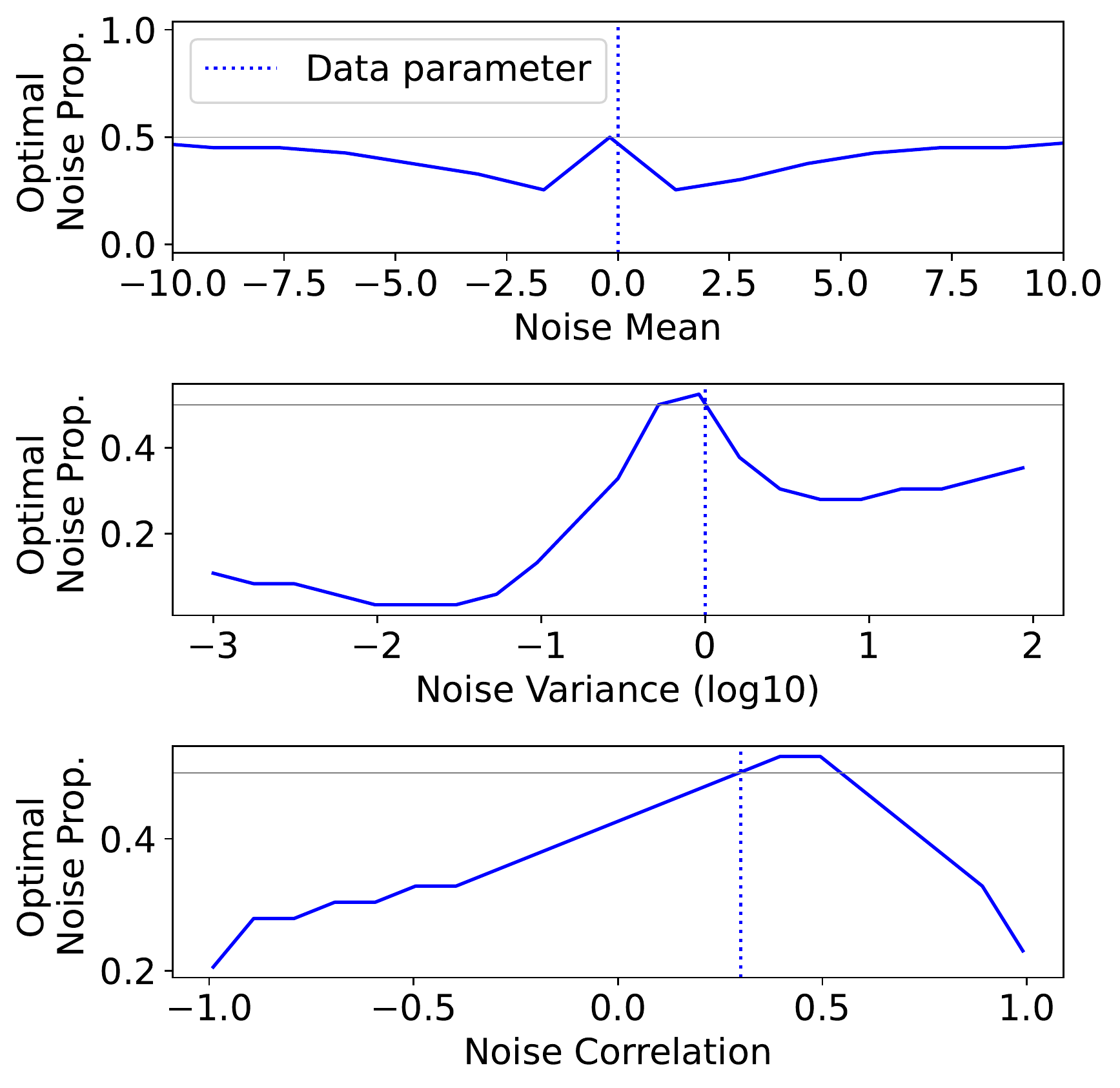}
\caption{Optimal noise proportion against the noise parameter. Top panel for model (a), Gaussian mean; Middle panel for model (b), Gaussian variance; Bottom panel for model (c), Gaussian correlation. \aapo{Parameter in the data distribution is given by the dotted blue line.}}
\label{fig:optimalpropvsval}
\end{figure}

\paragraph{Optimal noise proportion}
We proved in Section~\ref{sec:theoretical_results_parametric} that in the special case when the noise distribution is equal to the data distribution, then the optimal noise proportion is $50 \%$. Next we show empirically that the converse of this result does not hold: a noise proportion of $50\%$ does \textit{not}
ensure that the noise distribution equals the data's. A  counter-example was actually already found in Figure~\ref{fig:noiseval_vs_dataval}, where in the case of $\nu=1$, the optimal parameter for the noise was not equal to the parameter generating the data.  A closer look at this phenomenon is given by Figure~\ref{fig:optimalpropvsval} which shows the optimal noise proportion as a function of a Gaussian's parameter (mean, variance, or correlation). We see that while it is $50\%$ for when the data parameter is used for noise, it is typically less.

\subsection{Estimating Unnormalized Models}
\label{ssec:model_unnormalized}

\begin{figure}[!ht]
\centering
\begin{subfigure}[t]{0.48\textwidth}
    \centering
    \includegraphics[width=0.9\linewidth]{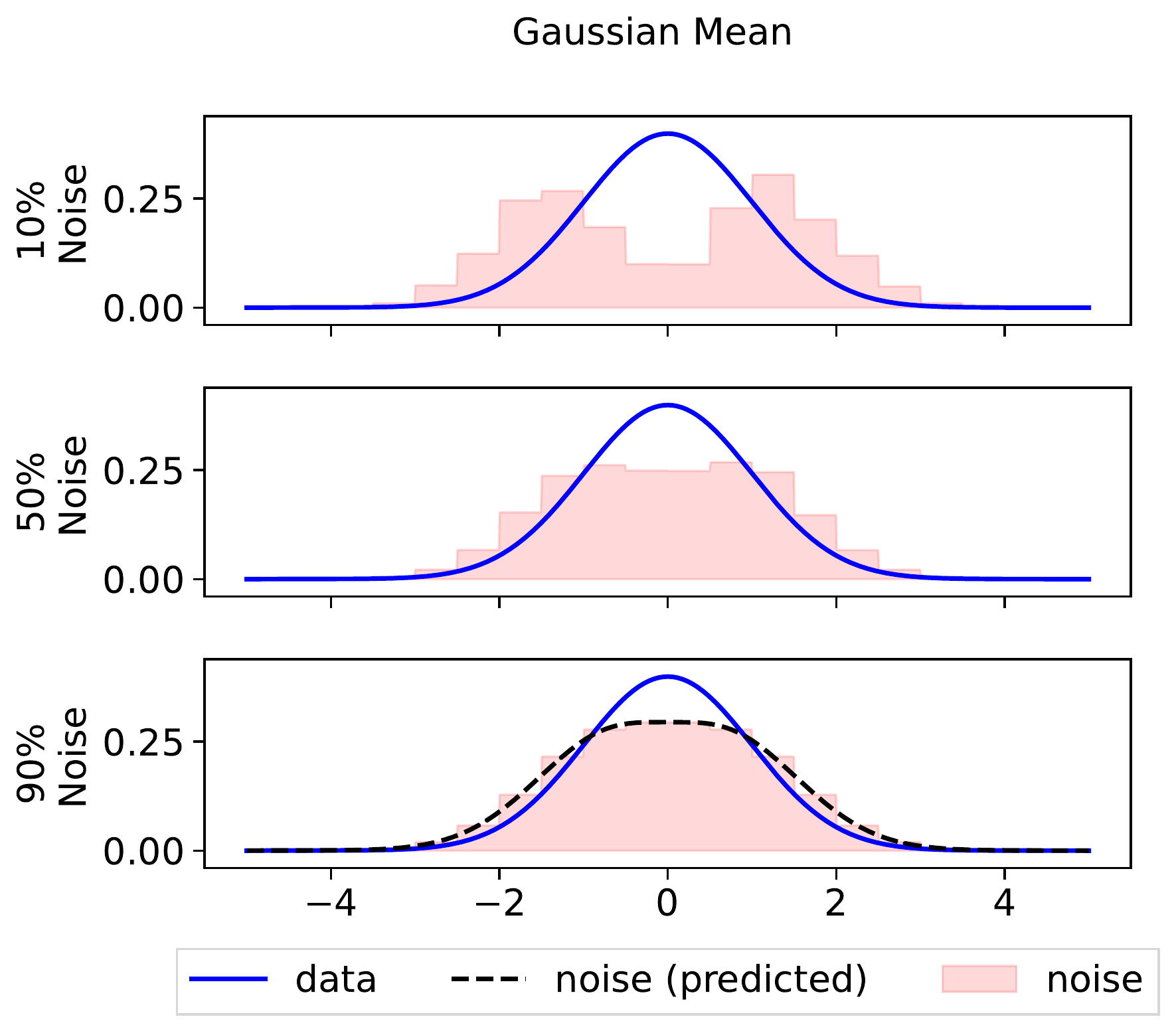} 
    \caption{Model a: Gaussian mean}
    \label{fig:optimalnoisemean_ebm}
\end{subfigure}
\hfill
\begin{subfigure}[t]{0.48\textwidth}
    \centering
    \includegraphics[width=0.9\linewidth]{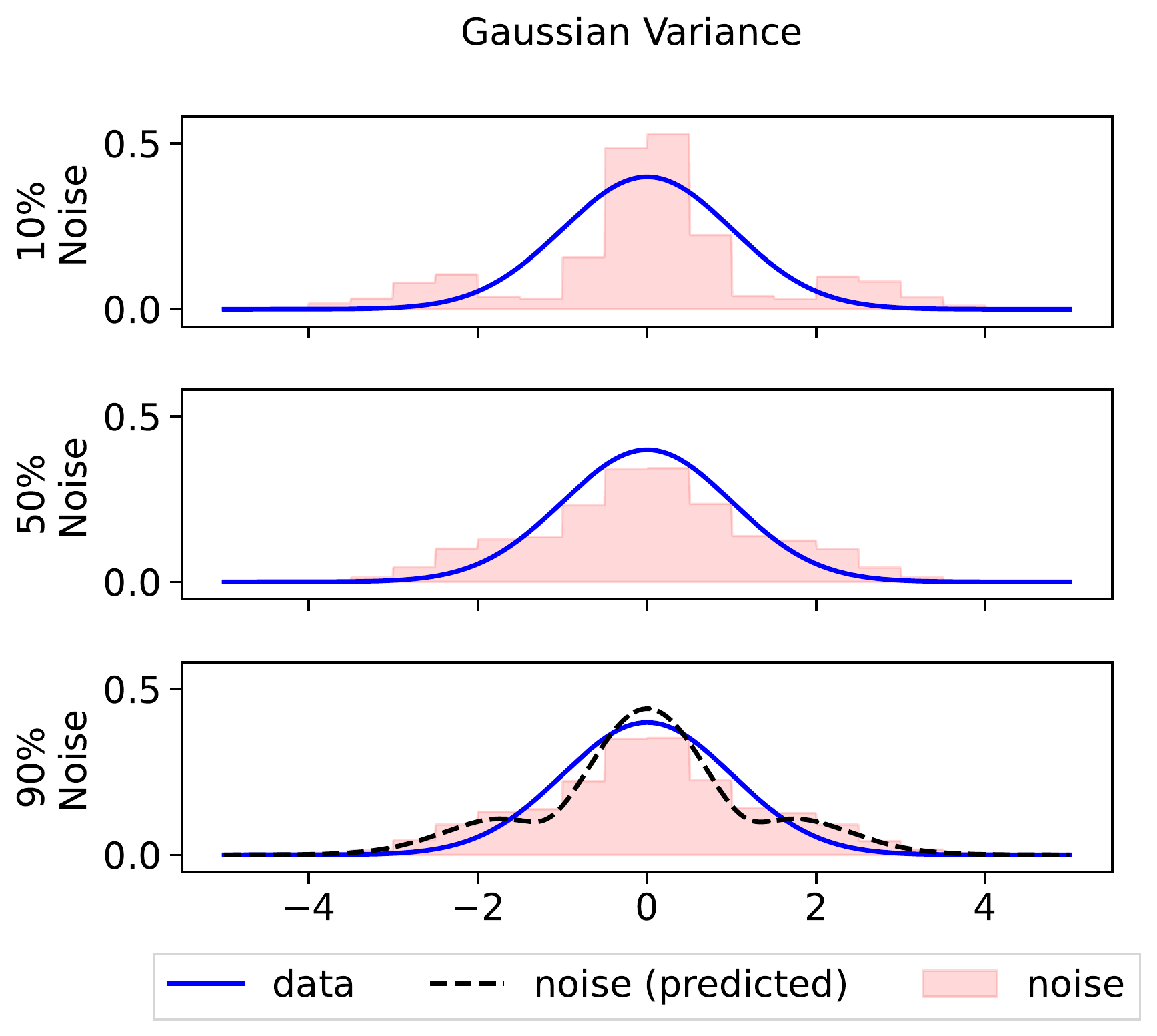}
    \caption{Model b: Gaussian variance}
    \label{fig:optimalnoisevar_ebm}
\end{subfigure}
\caption{Unconstrained (nonparametric histogram-based) optimal noise distributions when the data model is \textit{unnormalized}. Each row gives a different $\nu$ or noise proportion. The pink bars give the numerical approximations of the optimal noise distribution. The theoretical approximation of optimal noise is given by the dashed lines: the all-noise limit in the bottom panel, and the all-data limit in the top panel.}
\end{figure}

Figures~\ref{fig:optimalnoisemean_ebm} and~\ref{fig:optimalnoisevar_ebm} are obtained similarly to~\ref{fig:optimalnoisemean} and~\ref{fig:optimalnoisevar}, except that the log-normalization constant is estimated simultaneously to the original parameter (mean or variance). 
The effect of including the log-normalization constant as a parameter results in the optimal noise distribution resembling more the data distribution in the all noise limit, where our numerical results (in red) match our theoretical predictions (in black). Altogether, the optimal noise distribution seems to have a wider support when the log-normalization constant is included in the estimation. This is coherent with our theoretical results in section~\ref{sec:theoretical_results_general}, which 
state that the optimal noise must place its mass where the data distribution is high and varies most when $\theta^*$ and $c^* = \log Z^*$ change. Unlike the variation caused by the former $\nabla_{\vtheta} \log p(\vx; (\vtheta, c))$ which generally depends on $\vx$, the variation caused by the log-normalization constant $\nabla_{c} \log p(\vx; (\theta, c)) = -1$ is constant over the entire space. This suggests that including the log-normalization constant in the estimation causes the optimal noise distribution to have a wider support.

\section{Discussion}
\label{sec:discussion}

\subsection{Toward Statistical Efficiency}
In this paper, we have discussed choosing the noise distribution $\pn$ such that it minimizes the (asymptotic) variance of the NCE estimator, defined in~\eqref{eq:gnce_bregman}. We showed that in some limit cases, an optimal noise distribution can be built by reweighting the data distribution with some function of the Fisher score. 
Yet, even with such an optimal noise distribution, the variance does not reach the Cramer-Rao lower bound for the parameters of the energy. For normalized models, this is shown in Theorem~\ref{th:mse_gap_allnoise}
This means that the NCE estimator as it is currently defined, cannot achieve statistical efficiency. This has inspired recent modifications~\citep{uehara2018nce,uehara2020nceefficiency} to the estimator in order to further reduce its variance, which we next discuss.

\paragraph{Pre-estimating the noise distribution from the noise sample} 
Above, the noise distribution has been viewed as a design parameter, or hyperparameter. With a slight abuse of terminology, it could also be viewed as a \textit{nuisance parameter}~\citep{uehara2020nceefficiency}: it is not the main interest (estimating the data distribution) yet it affects the estimation.  This motivates a ``two-step M-estimation"~\citep{vandervaart2002semiparam,lok2021twostepestim}, which consists in estimating a nuisance parameter --- here the noise distribution --- from its sample before plugging it into the original estimator. 
This is in contrast to the original NCE estimator in~\eqref{eq:gnce_bregman} which is built using noise samples that are given as an \textit{empirical} noise distribution, while the objective uses the density of the \textit{true} noise distribution, leading to a mismatch. This idea was applied to the noise distribution in Importance Sampling~\citep{henmi2007isnoise} before making its way to NCE~\citep[Theorem 2]{uehara2018nce} where it led to a quantifiable decrease in the variance of the estimator~\citep[Eq. 4.30]{uehara2018nce}. For this new estimator, the optimal classification loss is no longer the JS classification loss, but the KL loss which generalizes Importance Sampling (Section~\ref{sec:partition_theoretical_results_parametric}). 

\paragraph{Correlating the data and noise samples} 
In general, the noise sampling mechanism is itself a source of variance and an implicit design choice of the estimator. 
In fact, independent data and noise samples can be suboptimal and a source of additional variance~\citep{pihlaja2010nce}. Using ``Common Random Numbers'~\citep{glasserman1992varreduction}, here correlating the noise and data samples, is a classic statistical method for variance reduction leading to ``control variates". This was applied to NCE in~\citep{uehara2020nceefficiency} by setting the noise sample equal to the data sample. Note that correlating the noise samples with the data sample was also considered for a different but related estimator, Conditional Noise Contrastive Estimation (CNCE)~\citep{ceylan2018conditionalnce}. 

\paragraph{Pre-estimating the noise distribution from the data sample} 
Both aforementionned methods --- pre-estimating the noise distribution and correlating its samples with the data --- are combined for NCE in~\citet{uehara2020nceefficiency}. They show that taking the noise sample equal to the data sample \textit{and} pre-estimating the noise distribution using the data sample, achieves the Cramer-Rao lower bound for estimating the energy parameter with NCE, under some conditions.
This comes with a strong assumption on the quality of the estimate of the noise distribution: its estimation error must must decrease quicker with the sample size than the data parameter's variance which is $O(1/T_d)$, so that it can be neglected in the asymptotic analysis~\footnote{\citet[Th.3]{uehara2020nceefficiency} use a non-parametric, higher-order kernel density estimator which achieves a rate of $O(1/T_d^4)$.}.
Intuitively, this can be understood as an instance of Rao-Blackwellisation: the original NCE estimator is conditioned on (a function of) $\pn$ which is pre-estimated using the data sample and is thus a sufficient statistic for (re)estimating the data model. 
Using these methods of variance-reduction, the optimal noise is no longer the data distribution corrected by the Fisher score, but the data distribution itself whose samples are already available. However, its density is not readily available: this is a chicken-and-egg problem, where the target (data distribution) cannot be plugged-in as the noise distribution because it is precisely what we are trying to estimate. 
Algorithmically, this implies that adapting the noise so that it finally targets the data distribution~\citep{gao2020fce} is statistically optimal at convergence \textit{if} the samples of noise and data are equal (not only in distribution, but obtained by the same ``seed"). If the samples are independent, there is evidence that setting the noise distribution equal to the data distribution, while not optimal, is good enough~\citep{gutmann2012nce} in the sense that it achieves a multiplicative factor of the Cramer-Rao lower bound. 

An interesting topic for future work would be to extend our analysis to such recent extensions of NCE which approach statistical efficiency.

\paragraph{Direct connection to MLE in terms of gradients} 
It has further been shown that when restricted to normalized models and setting the noise distribution equal to the data distribution, the NCE gradient becomes, in expectation, the MLE gradient~\citep{goodfellow2014gan,gutmann2022contrastivereview}. (This does not, however, guarantee efficiency since the variance of the NCE gradient could be larger.) Whenever the model is unnormalized, the NCE gradient in $\vtheta$ can be explicitly written as a reweighted version of the MLE gradient by the estimation error in the normalization~\citep[Eq. 45]{gutmann2022contrastivereview}. In other words, NCE approximates MLE when the noise is adapted to match the data model.
From a practical viewpoint, all the discussion above suggests that in practice, using algorithms where the noise targets the data distribution is likely to be optimal or very close. Nevertheless, some empirical work in NLP has proposed methods related to our theory that uses the Fisher score, as will be discussed below.

\subsection{Further related work}

\paragraph{Importance Sampling} 
A contribution of this paper is to show in more precise terms how NCE generalizes Importance Sampling (Section~\ref{sec:partition_theoretical_results_parametric}). Many earlier developments in NCE have already followed from an older literature of Importance Sampling, 
starting with the formulation of NCE as a variational problem~\citep[Eq. 2]{pihlaja2010nce}, 
its generalization using the $\alpha(\vx)$ reweighting function \citep{meng1996importancesamplingext} \citep[Equation 3.24]{uehara2018nce} covered in Section~\ref{ssec:theoretical_results_importance_sampling},
up to pre-estimating the noise distribution for variance reduction \citep{henmi2007isnoise} \citep[Theorem 2]{uehara2018nce}. 
Because of this connection, NCE also shares some problems with Importance Sampling, namely the choice of the noise distribution, and has consequently drawn from the same algorithmic developments proposed initially for Importance Sampling. These include adapting~\citep{gao2020fce} or annealing~\citep{rhodes2020bridgedre,choi2021bridgedre} the noise distribution to finally target the data distribution. 
It is therefore possible that our formalization of this connection may help draw further results from the Importance Sampling literature to improve NCE. 

\paragraph{Hard negatives} 
Our optimality results are coherent with empirical practice in Natural Language Processing (NLP), where ``hard negative" points are defined as points that yield high gradient amplitude of a classifier~\citep[Eqs.2-3]{kalantidis2020hardneg}. This is related to our optimal noise distributions, which reweigh the data density by the gradient of the classifier (Fisher score).
Our results may therefore amend the definition of~\citep{kalantidis2020hardneg}: hard negative (or noise) points should perhaps ideally be data points that yield a high gradient amplitude of the classifier.

\paragraph{Implications for Self-Supervised Learning} 
In more general terms, we have shown in this paper how statistical estimation theory can provide a framework to determine optimal design choices, or hyperparameters, for a pretext task in self-supervised learning. Specifically, Noise-Contrastive Estimation (NCE) as a method for estimating a density-ratio~\citep{sugiyama2012densityratiobook}
via classification, is a framework underlying recent advances in Nonlinear Independent Component Analysis (ICA)~\citep{hyvarinen2017pcl,hyvarinen2019gcl},Simulated-Based Inference (SBI)~\citep{durkan2020contrastivesbi,gutmann2022contrastivereview}, matrix completion or recommender systems~\citep{pellegrini2022amazondre}. We hope our analysis is a step towards a better understanding of how the sample efficiency of these methods depends on the choice of the noise distribution.

\section{Conclusion}
\label{ssec:conclusion}

 We studied the choice of optimal design parameters in Noise-Contrastive Estimation. We approached the problem from the viewpoint of statistical estimation theory, considering NCE as estimation of a parametric model. The asymptotic MSE, typically equivalent to the asymptotic variance, gives the gold standard for comparing estimators.  We started from the general framework using a family of Bregman divergences, and showed connections to Importance Sampling and its variants. However, from the the estimation theory viewpoint, the ordinary logistic NCE  is optimal in that family. Thus, the hyperparameters to be optimized, for a fixed computational budget, are essentially the noise distribution and the proportion of noise, the former being our focus here. 
 While optimizing the noise distribution seems intractable in the general case, it is easy to show empirically that, in stark contrast to what is often assumed, the optimal noise distribution is not the same as the data distribution (when the energy is estimated as well as the normalization), thus extending the analysis by \citet{pihlaja2010nce}. Our main theoretical results derive the optimal noise distribution in limit cases where either almost all observations to be classified are noise, or almost all observations are real data, or the noise distribution is an (infinitesimal) perturbation of the data distribution. The optimal noise distributions are different in these cases but have in common the point of emphasizing parts of the data space where the Fisher score function changes rapidly. 
 We hope these results will help improve the performance of NCE and related self-supervised methods in demanding applications.

\subsubsection*{Acknowledgements}

Numerical experiments were made possible thanks to the 
scientific Python ecosystem: 
Matplotlib~\citep{matplotlib}, 
Scikit-learn~\citep{scikit-learn}, 
Numpy~\citep{numpy}, 
Scipy~\citep{scipy} and PyTorch~\citep{pytorch}.


This work was supported by the French ANR-20-CHIA-0016 to Alexandre Gramfort. Aapo Hyv{\"a}rinen was supported by funding from the Academy of Finland and a Fellowship from CIFAR.

We are grateful to Michael Gutmann, Takeru Matsuda, Andrej Risteski and Frank Nielsen for interesting discussions.

\vskip 0.2in
\bibliography{paper}

\newpage

\appendix

\clearpage
\newpage

\section{Noise-Contrastive Estimation (NCE) and its generalizations}
\label{app:nce_generalizations}

\subsection{Bregman Classification Loss}
\label{app:ssec:nce_bregman_formulation}

\paragraph{Definition}
We here recall the necessary formalism for a Bregman divergence between two positive distributions $f, g$. A Bregman divergence is the sum of an entropy and cross-entropy
\begin{align*}
    D_{\phi}(f, g)
    &=
    H(f) 
    + 
    H(f, g)
\end{align*}
both of which are built off a convex function $\phi(x)$, as
\begin{align*}
    H(f) 
    &:=
    \int \phi(f) \, d\lambda(x)
    \\
    H(f, g) 
    &:=
    -\int
    \left(
    \phi(g) + \phi^{'}(g)(f - g) 
    \right)
    d \lambda(x)
    =
    -\left(
    H(g)
    +
    \nabla H(g) (f - g)
    \right)
    \enspace .
\end{align*}
where the dependencies of $f$ and $g$ on $x$ are dropped for notational convenience.
Geometrically, the Bregman divergence evaluates the ``convexity gap", between the graph of $\phi$ (entropy) at $f$, and its tangent line (cross-entropy) at $f$ from $g$. 
Perhaps the most famous Bregman divergence is the (generalized) Kullback-Leibler, achieved for $\phi(x) = x \log(x) - (1 + x) \log(1 + x)$. 
Minimizing the Bregman divergence with respect to $g$ amounts to minimizing the cross-entropy, equivalent to a scoring rule~\citep[Theorem 4.1]{ovcharov2018bregmanscore}.

\paragraph{Application to NCE}
In this manuscript, we compare the optimal and model classifiers by their respective density ratios, respectively $f = \frac{\pd}{\nu \pn}$ and $g = \frac{p_{\vbeta}}{\nu \pn}$. The Bregman divergence becomes
\begin{align*}
    D_{\phi} \left(
    \frac{\pd}{\nu \pn}
    , 
    \frac{p_{\vbeta}}{\nu \pn}
    \right)
    &=
    H \left( 
    \frac{\pd}{\nu \pn} 
    \right) 
    + 
    H \left( 
    \frac{\pd}{\nu \pn}
    ,
    \frac{p_{\vbeta}}{\nu \pn}
    \right) 
\end{align*}
when integration is performed with respect to the weighted noise $\lambda(\vx) = \nu \pn(\vx)$. The cross-entropy defines a classification loss 
\begin{align*}
    H \left( 
    \frac{\pd}{\nu \pn}
    ,
    \frac{p_{\vbeta}}{\nu \pn}
    \right) 
    &=
    \nu \E_n
    \bigg[
    S_0
    \bigg(
    \frac{p_\vbeta}{\nu \pn}
    \bigg)
    \bigg]
    -
    \E_d
    \bigg[
    S_1
    \bigg(
    \frac{p_\vbeta}{\nu \pn}
    \bigg)
    \bigg]
\end{align*}
after expanding, where for notational convenience we use $(S_0, S_1)$ which are two nonlinearities coupled by
\begin{alignat*}{2}
    S_0(x) = -\phi(x) + \phi'(x)x  
    \hspace{2em}
    S_1(x) = \phi'(x)
\end{alignat*}
\noindent or conversely
\begin{align*}
    \phi(x) &= S_1(x) x - S_0(x)
    \enspace .
\end{align*}
NCE minimizes that classification loss. Different values of $\phi$ define different classification losses. 
For example, $\phi(x) = x\log x - (1+x)\log(1+x) $ recovers the logistic loss. Another choice, $\phi(x) = (1- \sqrt{x})^2$, recovers the exponential loss. More generally, we name the classification losses as in~\citep[Section 2.2]{uehara2018nce}, based on terminology from the remaining entropy term
\begin{align*}
    H \left( 
    \frac{\pd}{\nu \pn} 
    \right) 
    :=
    \nu \E_n
    \left[
    \phi
    \big(
    \frac{\pd}{\nu \pn}
    \big)
    \right]
\end{align*}
The entropy is \textit{itself} a Czisar divergence between $\pd$ and the scaling density $\pn$ when $\nu = 1$~\citep{stummer2012scaledbregman}. For the entropy, choosing $\phi(x) = x\log x - (1+x)\log(1+x)$ defines the Jensen-Shannon (JS) divergence between $\pd$ and $\pn$. Likewise, choosing $\phi(x) = (1- \sqrt{x})^2$ defines  defines the squared Hellinger ($H^2$) distance. Two additional choices of $\phi$ will be of interest to us in this manuscript: $\phi(x) = x \log x - x$ defining the Kullback-Leibler (KL) divergence and $\phi(x) = -1 -\log(x)$ defining the reverse-KL divergence.

\paragraph{Connection to GANs}
If this cross-entropy is brought to zero, this leaves the entropy which is a Czisar divergence between $\pd$ and the scaling density $\pn$~\citep{stummer2012scaledbregman} when $\nu = 1$. Minimizing then the entropy provides a second way to estimate data distribution~\citep{nguyen2010ratioestim}, used in Generative Adversarial Networks (GANs) ~\citep{goodfellow2014gan,nowozin2016fgan}.

\section{Measuring the Estimation Error of NCE}
\label{app:sec:estimation_error_gnce}

For the Bregman generalization of NCE, where the classification loss (indexed by $\phi$) is itself a hyperparameter, the asymptotic variance matrix is~\citep{pihlaja2010nce}
\begin{align*}
    \mSigma 
    &= 
    \mI_{w}^{-1} \big(
    \mI_{v} 
    -
    (1 + \frac{1}{\nu})
    \vm_{w} \vm_{w}^\top
    \big)
    \mI_{w}^{-1}   
    \numberthis
    \label{eq:asympmsegnce}
\end{align*}
where $\vm_w(\vbeta^*)$, $\mI_w(\vbeta^*)$ and $\mI_v(\vbeta^*)$ are a generalized score mean and covariances, reweighed by
$w(\vx) = \frac{\pd}{\nu \pn}(\vx) \phi^{''}(\frac{\pd}{\nu \pn}(\vx))$
and
$v(\vx) = w(\vx)^2 P(Y=0 | \vx)^{-1}$
:
\begin{align*}
    \vm_w(\vbeta^*)
    &=
    \E_d
    \,
    w(\vx)
    \left(
    \nabla_{\vbeta} F(\vx, \vbeta^*)
    \right)
    \\
    \mI_w(\vbeta^*) & = 
    \E_d 
    w(\vx)
    \left(
    \nabla_{\vbeta} F(\vx, \vbeta^*)    \,
    \nabla_{\vbeta} F(\vx, \vbeta^*)^\top 
    \right)
    \\
    \mI_v(\vbeta^*) & = 
    \E_d 
    w(\vx)
    \left(
    \nabla_{\vbeta} F(\vx, \vbeta^*)    \,
    \nabla_{\vbeta} F(\vx, \vbeta^*)^\top 
    \right)
\end{align*}

\paragraph{Parametric Estimation Error} 
The parametric estimation error is therefore
\begin{align*}
    \mathrm{MSE}_{\hat{\vbeta}}(\pn) 
    &=
    \frac{\nu + 1}{T} \mathrm{tr}
    (
    \mSigma
    ) 
\end{align*}
\paragraph{Non-parametric Estimation Error}
The non-parametric estimation error is defined as \omar{the mean Kullback-Leibler divergence (MKL) between the data distribution and the estimated model} 
\begin{align*}
    \omar{
    \mathrm{MKL}_{\hat{\vbeta}}(\pn) 
    }
    &=
    \mathbb{E}\big[ \mathcal{D}_{\mathrm{KL}}(p_d, p_{\hat{\vbeta}}) \big]
\end{align*}
where the generalized (Bregman) Kullback-Leibler divergence is used:
\begin{align*}
    \mathcal{D}_{\mathrm{KL}}(f, g) 
    :=
    \int \left(
    f(\vx) \log(f(\vx)/g(\vx)) - f(\vx) + g(\vx) 
    \right) d\vx
    \enspace .
\end{align*}
We can specify this error, by using the Taylor expansion of the estimated $\hat{\vbeta}$ near optimality, given in~\cite{gutmann2012nce}: 
\begin{align*}
    \hat{\vbeta} - \vbeta^*
    & =
    \vz
    + 
    O(\|\hat{\vbeta} - \vbeta^*\|^2)
    \numberthis
    \label{eq:paramerror}
\end{align*}
where 
$z \sim \mathcal{N}(0, \frac{1}{T_d}\mSigma)$ 
and $\mSigma$ is the asymptotic variance matrix. We can similarly take the Taylor expansion of the generalized KL divergence with respect to its second argument, near optimality:
\begin{align*}
    J(\hat{\vbeta})
    & :=
    \mathcal{D}_{\mathrm{KL}}(p_d, p_{\hat{\vbeta}}) \\
    & = 
    J(\vbeta^*) 
    + 
    \langle \nabla_{\vbeta}J(\vbeta^*), \hat{\vbeta} - \vbeta^* \rangle
    + 
    \frac{1}{2}
    \langle (\hat{\vbeta} - \vbeta^*), \nabla^2_{\vbeta}J(\vbeta^*) \, (\hat{\vbeta} - \vbeta^*) \rangle
    +
    O(\| \hat{\vbeta} - \vbeta^* \|^3) \\
    & = 
    J(\vbeta^*) 
    + 
    \langle \nabla_{\vbeta}J(\vbeta^*), \hat{\vbeta} - \vbeta^*) \rangle
    + 
    \frac{1}{2}
    \| \hat{\vbeta} - \vbeta^* \|^2_{\nabla^2_{\vbeta}J(\vbeta^*)}
    +
    O(\| \hat{\vbeta} - \vbeta^* \|^3) 
\end{align*}
Note that some simplifications occur:
\begin{itemize}
    \item[] $J(\vbeta^*) = \mathcal{D}_{\mathrm{KL}}(p_{\vbeta^*}, p_{\vbeta^*}) = 0$ 

    \item[] $\nabla_{\vbeta}J(\vbeta^*) = 0$
    as the gradient the generalized KL divergence with respect to its second argument is null.
    
    \item[] $\nabla_{\vbeta}^2J(\vbeta^*) = \mI$
\end{itemize}
Plugging in the estimation error from~\eqref{eq:paramerror} into the distribution error yields:
\begin{align*}
    J(\hat{\vbeta}) 
    & =
    \frac{1}{2}
    \bigg\| 
    \vz + O(\| \hat{\vbeta} - \vbeta^* \|^2) \bigg\|_{\mI}^2
    +
    O(\| \hat{\vbeta} - \vbeta^* \|^3) \\
    & =
    \frac{1}{2}
    \bigg(
    \| \vz \|_{\mI}^2 
    + 
    2
    <\vz, O(\| \vbeta - \vbeta^*\|^2)>_{\mI}
    + 
    \big\| O(\| \vbeta - \vbeta^*\|^2) \big\|_{\mI}^2 
    \bigg) + O(\| \hat{\vbeta} - \vbeta^* \|^3) \\
    & = 
    \frac{1}{2} \| \vz \|_{\mI}^2 
    +
    O(\| \hat{\vbeta} - \vbeta^* \|^2)
\end{align*}
by truncating the Taylor expansion to the first order. Hence up to the first order, the expectation yields:
\begin{align*}
    \mathbb{E}\big[ \mathcal{D}_{\mathrm{KL}}(p_d, p_{\hat{\vbeta}}) \big] 
    & =
    \frac{1}{2}
    \mathbb{E}\big[ \| \vz \|_{\mI}^2 \big]
    =
    \frac{1}{2}
    \mathbb{E}\big[ \vz^\top \mI \vz \big]
    =
    \frac{1}{2}
    \mathbb{E}\big[ \mathrm{tr}(\vz^\top \mI \vz) \big] 
    =
    \frac{1}{2}
    \mathbb{E}\big[ \mathrm{tr}(\mI \vz \vz^\top) \big] \\
    & =
    \frac{1}{2}
    \mathrm{tr}( \mI \mathbb{E}[\vz \vz^\top])
    =
    \frac{1}{2}
    \mathrm{tr}( \mI \mathrm{Var}[\vz])
    =
    \frac{1}{2 T_d}
    \mathrm{tr}( \mI \mSigma)
\end{align*}
Note that this is a general and known result which is applicable beyond the KL divergence: for any divergence, the 0th order term is null as it measures the divergence between the data distribution and itself, the 1st order term is null in expectation if the estimator $\hat{\vbeta}$ is asymptotically unbiased, which leaves an expected error given by the 2nd-order term $\frac{1}{2T_d}\mathrm{tr}(\nabla^2 J \,\, \mSigma)$ where J is the chosen divergence as a function of its second argument. Essentially, one would replace the generalized Fisher Information matrix $\mI$, which is the Hessian for the generalized KL divergence, by the Hessian for another Bregman divergence.

Finding the optimal noise that minimizes the distribution error means minimizing $\frac{1}{T_d} \mathrm{tr}(\mSigma \mI)$. One can contrast that with the optimal noise that minimizes the parameter estimation error (asymptotic variance) $\frac{1}{T_d} \mathrm{tr}(\mSigma)$. 

\subsection{Parametric Estimation Error for the normalization}
\label{app:ssec:estim_error_partition}

We will here specify the estimation error for different estimators in the NCE family, whenever the data model is parameterized only by the partition $Z$. The formula we will use throughout is
\begin{align*}
    \mathrm{MSE}(\pn) 
    &=
    \frac{1 + \nu}{T}
    Z^{*^2}
    \bigg(
    \frac{\E_d \big[
    w^2 \frac{\nu \pn + \pd}{\nu \pn}
    \big]}
    {\E_d[w]^2}
    -
    (1 + \frac{1}{\nu})
    \bigg)
\end{align*}
where the MSE depends on the classification loss identified by $\phi(x)$ via the reweighting
\begin{align*}
    w(\vx) 
    :=
    \frac{\pd}{\nu \pn}(\vx) \phi^{''}\big( 
    \frac{\pd}{\nu \pn}(\vx) \big)
    \enspace .
\end{align*}
We directly obtain the MSE for different classification losses, by plugging in different convex functions $\phi(x)$:
\begin{center}
    \begin{tabular}{ ccccc } 
    \toprule
    Name 
    & 
    Loss identified by $\phi(\vx)$
    & 
    $\phi^{''}(\vx)$
    & 
    Reweighting $w(\vx)$
    &
    MSE
    \\
    \midrule
    KL
    & 
    $x \log x$
    & 
    $\frac{1}{x}$
    &
    $1$
    &
    $\frac{1 + \nu}{\nu T} Z^{*^2} \mathcal{D}_{\chi^2}(\pd, \pn)$
    \\ 
    \midrule
    revKL
    & 
    $- \log x$
    & 
    $\frac{1}{x^2}$
    &
    $\frac{\nu \pn(\vx)}{\pd(\vx)}$
    &
    $\frac{1 + \nu}{T} Z^{*^2} \mathcal{D}_{\chi^2}(\pn, \pd)$
    \\ 
    \midrule
    JS
    & 
    $x \log x - (1 + x) \log(\frac{1 + x}{2})$
    & 
    $\frac{1}{x (1 + x)}$
    &
    $\frac{\nu \pn}{\nu \pn + \pd}$
    &
    $\frac{(1 + \nu)^2}{\nu T} Z^{*^2}
    \frac{\mathcal{D}_{\mathrm{HM}}(\pd, \pn)}{1 - \mathcal{D}_{\mathrm{HM}}(\pd, \pn)}$
    \\ 
    \midrule
    $H^2$
    & 
    $(1 - \sqrt{x})^2$
    & 
    $\frac{1}{2}x^{-3/2}$
    &
    $\frac{1}{2} \sqrt{\frac{\nu \pn(\vx)}{\pd(\vx)}}$
    &
    $\frac{(1 + \nu)^2}{\nu T} Z^{*^2} \frac{\mathcal{D}_{H}(\pd, \pn)^2}{1 - \mathcal{D}_{H}(\pd, \pn)^2}$
    \\
    \bottomrule
    \end{tabular}
\end{center}
The MSE is consistently expressed in terms of divergences between the data and noise distributions, namely: 
\begin{itemize}

  \item[] $\mathcal{D}_{\chi^2}(\pd, \pn) := \big( \int \frac{\pd^2}{\pn} \big) - 1$ is the chi-squared divergence
  
  \item[] $D_{H}(\pd, \pn) := \sqrt{1 - \big( \int \sqrt{\pd \pn} \big)} \in [0, 1]$ is the Hellinger distance

  \item[] $\mathcal{D}_{\mathrm{HM}}(\pd, \pn) := 1 - \int \big( \pi \pd^{-1} + (1 - \pi) \pn^{-1} \big)^{-1} = 1 - \frac{1}{\pi} \E_\pd \frac{\pi \pn}{(1 - \pi) \pd + \pi \pn}
    \in [0, 1]$ 
    \\
    is the harmonic divergence with weight $\pi \in [0, 1]$.
    \\
     Here, the weight $\pi = P(Y=0) = \frac{T_n}{T} = \frac{\nu}{1 + \nu}$. 

\end{itemize}
Note that the MSE for the Squared-Hellinger Classification loss is similar to the MSE for the Jensen-Shannon Classification loss: both are proportional to $\frac{d}{1 - d}$, where $d \in [0, 1]$ is a divergence between the data and noise distributions. 

%

\newpage

\section{Finding the optimal classification contrast}
\label{app:sec:gnce_optimal_noise_distribution}

Having established that the optimal loss (indexed by $\phi$) for NCE is the logistic classification loss, we now look for the optimal noise distribution $\pn$. The goal is to optimize the $\mathrm{MSE}(T, \nu, \pn)$ with respect to the noise distribution $\pn$, where
\begin{align*}
    \mathrm{MSE}(T, \nu, \pn) 
    &=
    \frac{\nu + 1}{T} \mathrm{tr}
    (
    \mI_w^{-1} - \frac{\nu + 1}{\nu} (\mI_w^{-1} \vm_w \vm_w^\top \mI_w^{-1})
    )
\end{align*}
where $\vm_w$ and $\mI_w$ are a generalized score mean and covariance, reweighted by misclassified points.  These integrals depend non-linearly on $\pn$ via
the reweighting function $w(\vx) = P(Y=0 | \vx) = \frac{\nu \pn(\vx)}{\pd(\vx) + \nu \pn(\vx)}$
\begin{alignat*}{2}
    \vm_w(\vbeta^*) 
    = 
    \int
    \fisherscore[\vx]
    \,
    w(\vx) d\vx
    \hspace{1em}
    \mI_w(\vbeta^*) 
    = 
    \int
    \fisherscore[\vx]
    \fisherscore[\vx]^\top 
    w(\vx) d\vx
    \enspace .
\end{alignat*}
and are additionally inversed and multiplied, which makes the dependency on $\pn$ harder to study.

\subsection{Intractability of the 1D Gaussian case}
\label{ssec:intractability}

Suppose a simple model, where the data distribution $\pd$ is a one-dimensional, zero-mean and unit-variance Gaussian distribution. It is normalized, so the normalization constant is known, not estimated. The model and noise distributions are of the same family, parameterized by mean and/or variance (we write these together in one model):
\begin{alignat*}{2}
    p(x; (\mu, \alpha)) 
    = 
    \frac{1}{\sqrt{2\pi\alpha}}e^{-\frac{1}{2}\frac{(x-\mu)^2}{\alpha}} 
    \hspace{2em}
    \pn(x, (\pi, \gamma)) = 
    \frac{1}{\sqrt{2\pi\gamma}}e^{-\frac{1}{2}\frac{(x-\pi)^2}{\gamma}} \qquad x \in \R
\end{alignat*}
It follows that the 2D score is written as:
\begin{align*}
    \nabla_{\mu, \alpha} F(x, (\mu^*, \alpha^*)) 
    = 
    \begin{pmatrix}
        \partial_\mu \log p(x; (\mu, \alpha)) 
        \\
        \partial_\alpha \log p(x; (\mu, \alpha))
    \end{pmatrix} \bigg|_{\mu=0,\alpha=1}
    =
    \begin{pmatrix}
        x 
        \\
        -1+ x^2
    \end{pmatrix}
\end{align*}
and its ``pointwise covariance" is:
\begin{align*}
    \nabla_{\mu, \alpha} F(x, (\mu^*, \alpha^*))
    \nabla_{\mu, \alpha} F(x, (\mu^*, \alpha^*))^\top
    =
    \begin{pmatrix}
        x^2 & -x + x^3 \\
        -x + x^3 & x^4 - x^2 + 1
    \end{pmatrix}
\end{align*}
In the following, we consider estimation of variance only, i.e.\ only the second term $\vm_2$ of the generalized score mean $\vm$ and the second diagonal term $\mI_{22}$ of the score variance matrix $\mI$,  as they intervene in the MSE formula for Noise-Contrastive Estimation:
\begin{align*}
    \vm_2 
    & = 
    \int
    \partial_\alpha F(x; (\mu, \alpha))
    \,
    \frac{\nu \pn(x; (\pi, \gamma))}{\pd(x) + \nu \pn(x; (\pi, \gamma))} dx
    \\
    & =
    -\frac{1}{2\sqrt{2\pi}} 
    \int 
    \left( e^{\frac{-x^2}{2}} \frac{1}{1 + \frac{1}{\nu}\sqrt{\gamma} e^{\frac{-x^2}{2}({1} - \frac{1}{\gamma})}} \right)dx 
     + 
    \frac{1}{2\sqrt{2\pi}} 
    \int 
    x^2
    \left( e^{\frac{-x^2}{2}} \frac{1}{1 + \frac{1}{\nu}\sqrt{\gamma} e^{\frac{-x^2}{2}(1 - \frac{1}{\gamma})}} \right)dx  
\end{align*}
and 
\begin{align*}
    \mI_{22}
    & = 
    \int
    \partial_\alpha F(x; (\mu, \alpha))^2
    \frac{\nu \pn(x; (\pi, \gamma))}{\pd(x) + \nu \pn(x; (\pi, \gamma))} dx
    =
    \frac{1}{4\sqrt{2\pi}} 
    \int 
    x^4
    \left( e^{\frac{-x^4}{2}} \frac{1}{1 + \frac{1}{\nu}\gamma^{\frac{1}{2}} e^{\frac{-x^2}{2}(1 - \frac{1}{\gamma})}} \right) dx 
    \\
     & - 
    \frac{1}{2\alpha^3\sqrt{2\pi}} 
    \int 
    x^2
    \left( e^{\frac{-x^2}{2}} \frac{1}{1 + \frac{1}{\nu}\gamma^{\frac{1}{2}} e^{\frac{-x^2}{2}(1 - \frac{1}{\gamma})}} \right) dx
     +
    \frac{1}{4\sqrt{2\pi}} 
    \int 
    \left( e^{\frac{-x^2}{2}} \frac{1}{1 + \frac{1}{\nu}\gamma^{\frac{1}{2}} e^{\frac{-x^2}{2}(1 - \frac{1}{\gamma})}} \right)dx  
\end{align*}
We see that even in a simple 1D Gaussian setting,  evaluating the asymptotic MSE of the Noise-Contrastive Estimator is intractable in closed-form, given the integrals in $\mI_{22}$, where the integrand includes the product of a Gaussian density with the logistic function compounded by the Gaussian density, further multiplied by monomials. While here we considered the case of variance, the intractability is seen even in the case of the mean.
Optimizing the asymptotic MSE 
with respect to $\gamma$ and $\pi$ (noise distribution) or $\nu$ (identifiable to the noise proportion)
yields similarly intractable integrals.

We resort to visualizations of the MSE landscape of the NCE estimator, when the noise is constrained within a parametric family containing the data. 
\begin{figure}[!ht]
\centering
\includegraphics[width=\columnwidth]{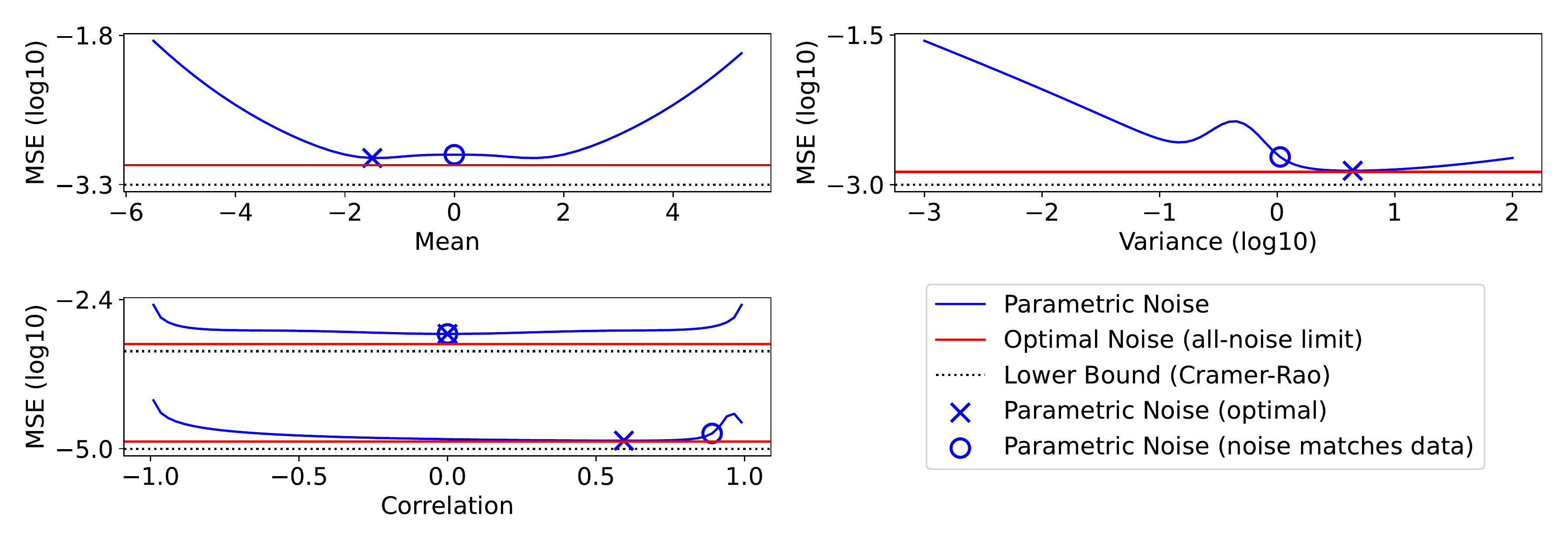}
\caption{MSE vs. the noise parameter. Top left panel for model (i), Gaussian mean; Top right panel for model (ii), Gaussian variance; Bottom left for model (iii), Gaussian correlation.
}
\label{fig:msevsnoiseval}
\end{figure} 
Namely, we draw attention to the two local minima symmetrically placed to the left and to the right of the Gaussian mean. This corroborates the indeterminacies observed in this paper (Conjecture on limit of zero noise), as to where the optimal noise should place its mass for this estimation problem. 

\subsection{Motivating the all noise and all data limits}
\label{app:ssec:limits_motivation}

The intractability of the MSE when the model being estimated is a simple, one-dimensional Gaussian, highlights the need for further simplifications.
The noise distribution which we are optimizing intervenes in the MSE only via the weighting function $w(\vx)$. Our general strategy will therefore be to consider limit cases in which the reweighting simplifies, namely the limits of all noise ($\nu \rightarrow \infty$) or all data ($\nu \rightarrow 0$) points.

Let us denote by $r = p_{\vbeta} / \nu \pn$ the density ratio. In the limit of all noise ($\nu \rightarrow \infty$, or $r \rightarrow 0$), the generator $\phi_{\mathrm{JS}}(r)$ of the JS classification loss tends to the generator $\phi_{\mathrm{KL}}(r)$ of the KL classification loss, modulo an affine function in $r$.
\begin{align*}
    \phi_{\mathrm{JS}}(r)
    & :=
    r \log r - (1 + r) \log \big( \frac{1 + r}{2} \big)
    =
    r \log r + r(\log 2 - 1) + \log(2) + \mathrm{O}_{\nu \rightarrow \infty}(\nu^{-2})
    \\
    & = 
    \phi_{\mathrm{KL}}(r) + r(\log 2 - 1) + \log(2) + \mathrm{O}_{\nu \rightarrow \infty}(\nu^{-2})
\end{align*}
Similarly, in the limit of all data ($\nu \rightarrow 0$, or $r \rightarrow \infty$), the generator $\phi_{\mathrm{JS}}(r)$ of the JS classification loss tends to the generator $\phi_{\mathrm{revKL}}(r)$ of the reverse KL classification loss, modulo an affine function in $r$.
\begin{align*}
    \phi_{\mathrm{JS}}(r)
    & :=
    r \log r - (1 + r) \log \big( \frac{1 + r}{2} \big)
    =
    -\log r + r \log 2 - 1 + \log 2 + \mathrm{O}_{\nu \rightarrow 0}(\nu^{1})
    \\
    & = 
    \phi_{\mathrm{revKL}}(r) + r \log 2 - 1 + \log 2 + \mathrm{O}_{\nu \rightarrow 0}(\nu^{1})
\end{align*}
Because the generating function $\phi$ of a Bregman divergence is defined modulo an affine function, we may interpret what happens in the limits of all noise and all data points: the JS classification loss tends to the KL and reverse KL classification losses, respectively. 

In the following, our proof structure will be to:
\begin{enumerate}
    \item Perform a Taylor expansion of the reweighting $w(\vx)$ in the $\nu \rightarrow 0$ or $\nu \rightarrow \infty$ limit
    \item Plug into the integrals $\vm_w$, $\mI_w$ and evaluate them (up to a certain order)
    \item Perform a Taylor expansion of  $\mI_w^{-1}$ (up to a certain order)
    \item Evaluate the $\mathrm{MSE}_{\mathrm{NCE}}$ (up to a certain order) \\
    \item Optimize the $\mathrm{MSE}_{\mathrm{NCE}}$ w.r.t. $\pn$
    \item Compute the MSE gaps at optimality
\end{enumerate}

\subsection{Limit of all noise}
\label{app:ssec:proof_mse_allnoise_parametric}

We here prove Theorem~\ref{th:allnoisebestmse}. 
We start with a change of variables $\gamma = \frac{1}{\nu} \rightarrow 0$ to bring us to a zero-limit. 
The MSE in terms of our new variable $\gamma = \frac{1}{\nu}$ can be written as:
\begin{alignat*}{2}
    &\mathrm{MSE}_{\mathrm{NCE}}(T, \gamma, \pn) 
    &&= 
    \frac{\gamma + 1}{\gamma T} \mathrm{tr}
    (\mI_w^{-1}) - \frac{(\gamma + 1)^2}{T \gamma} \mathrm{tr}(\mI_w^{-1} \vm \vm^\top \mI_w^{-1}) 
    \\
    &=
    (\gamma^{-1} T^{-1} + \gamma^0 
    && T^{-1}) \mathrm{tr}
    (\mI_w^{-1}) 
    - 
    (\gamma^{-1}T^{-1} + \gamma^0 2T^{-1} + \gamma^1 T^{-1}) 
    \mathrm{tr}(\mI_w^{-1} \vm_w \vm_w^\top \mI_w^{-1})
\end{alignat*}
Given the term up until $\gamma^{-1}$ in the MSE, we will use Taylor expansions up to order 2 throughout the proof, in anticipation that the MSE will be expanded until order 1.
\begin{enumerate}
    \item Taylor expansion of the reweighting 
    \begin{align*}
        w(\vx)
        = 
        \frac{\nu \pn(\vx)}{\pd(\vx) + \nu \pn(\vx)}
        = 
        \frac{1}{1 + \gamma \frac{\pd}{\pn}(\vx)}
        = 
        1 - \gamma \frac{\pd}{\pn}(\vx) + \gamma^2 \frac{\pd^2}{\pn^2}(\vx) + \circ(\gamma^2)
    \end{align*}
    \item Evaluating the integrals $\vm_w$, $\mI_w$
    \begin{align*}
        \vm_w 
        & = 
        \int \nabla_{\vbeta} F(\vx, \vbeta^*) 
        w(\vx) \pd(\vx) d\vx 
        \\
        & = 
        \int \nabla_{\vbeta} F(\vx, \vbeta^*)\pd(\vx)
        \bigg( 1 - \gamma \frac{\pd}{\pn}(\vx) + \gamma^2 \frac{\pd^2}{\pn^2}(\vx) + \circ(\gamma^2) \bigg) 
        d\vx \\     
        & = 
        \vm - \gamma \va + \gamma^2 \vb + \circ(\gamma^2) 
        \numberthis
    \end{align*}
    where $\vm$ is the Fisher-score mean of the (possibly unnormalized) model and we use shorthand notations $a$ and $b$ for the remaining integrals:
    \begin{align*}
        \vm
        = 
        \int 
        \nabla_{\vbeta} F(\vx, \vbeta^*)
        \pd(\vx)
        d\vx 
    \end{align*}
    \vspace{-1.5em}
    \begin{alignat*}{2}
        \va
        = 
        \int \nabla_{\vbeta} F(\vx, \vbeta^*)\frac{\pd^2}{\pn}(\vx)
        d\vx
        \hspace{1em}
        \vb
        = 
        \int \nabla_{\vbeta} F(\vx, \vbeta^*)\frac{\pd^3}{\pn^2}(\vx)
        d\vx \enspace .
    \end{alignat*}
    Similarly,
    \begin{align*}
        \mI_w 
        & = 
        \int \nabla_{\vbeta} F(\vx, \vbeta^*)\nabla_{\vbeta} F(\vx, \vbeta^*)^\top
        w(\vx) \pd(\vx) d\vx 
        \\
        & = 
        \int \nabla_{\vbeta} F(\vx, \vbeta^*)\nabla_{\vbeta} F(\vx, \vbeta^*)^\top \pd(\vx)
        \bigg( 1 - \gamma \frac{\pd}{\pn}(\vx) + \gamma^2 \frac{\pd^2}{\pn^2}(\vx) + \\
        & \circ(\gamma^2) \bigg) 
        d\vx    
        = 
        \mI - \gamma \mA + \gamma^2 \mB + \circ(\gamma^2)
    \end{align*}
    where the Fisher-score covariance (Fisher information) is $\mI$ and we use shorthand notations $A$ and $B$ for the remaining integrals:
    \begin{align*}
        \mI
        = 
        \int \nabla_{\vbeta} F(\vx, \vbeta^*)\nabla_{\vbeta} F(\vx, \vbeta^*)^\top \pd(\vx)
        d\vx 
    \end{align*}
    \begin{alignat*}{2}
        \mA
        = 
        \int \nabla_{\vbeta} F(\vx, \vbeta^*)\nabla_{\vbeta} F(\vx, \vbeta^*)^\top \frac{\pd^2(\vx)}{\pn(\vx)}
        d\vx 
        \hspace{0.5em}
        \mB
        = 
        \int \nabla_{\vbeta} F(\vx, \vbeta^*)\nabla_{\vbeta} F(\vx, \vbeta^*)^\top \frac{\pd^3(\vx)}{\pn^2(\vx)}
        d\vx \enspace .
    \end{alignat*}
    
    \item Taylor expansion of $\mI_w^{-1}$
    \begin{align*}
        \mI_w^{-1} 
        & = 
        \bigg(
        \mI - \gamma \mA + \gamma^2 \mB + \circ(\gamma^2) 
        \bigg)^{-1}
        = 
        \bigg(
        \mI (\textbf{Id} - \gamma \mI^{-1} \mA + \gamma^2 \mI^{-1} \mB)
        + \circ(\gamma^2) 
        \bigg)^{-1} \\
        & = 
        \mI^{-1}
        \bigg(
        \textbf{Id} - \gamma \mI^{-1} \mA + \gamma^2 \mI^{-1} \mB \bigg)^{-1}
        + \circ(\gamma^2) \\
        & = 
        \mI^{-1}
        \bigg(
        \textbf{Id} + \gamma \mI^{-1} \mA + \gamma^2 ((\mI^{-1} \mA)^2 - \mI^{-1} \mB) + \circ(\gamma^2)
        \bigg)
        + \circ(\gamma^2) \\
        & = 
        \mI^{-1} + \gamma \mI^{-2} \mA + \gamma^2 (\mI^{-1}(\mI^{-1} \mA)^2 - \mI^{-2} \mB) + \circ(\gamma^2)
    \numberthis
    \label{eq:taylorinverseIallnoise}
    \end{align*}
    
    \item Evaluating the $\mathrm{MSE}_{\mathrm{NCE}}$
    \begin{align*}
        \mathrm{MSE}_{\mathrm{NCE}} 
        =
        \frac{\gamma + 1}{\gamma T} \mathrm{tr}
        (\mI_w^{-1}) - \frac{(\gamma + 1)^2}{T \gamma} \mathrm{tr}(\mI_w^{-1} \vm_w \vm_w^\top \mI_w^{-1}) 
    \end{align*}
    We start by computing the second term of the MSE without the trace. Plugging in the Taylor expansions of $\mI_w^{-1}$ and $\vm_w$ and retaining only terms up to the second order yields
    \begin{align*}
        \mI_w^{-1} \vm_w \vm_w^\top \mI_w^{-1}
        & = 
        \mI^{-1} \vm \vm^\top \mI^{-1}
        \gamma^2 (
        \mI^{-1} \va \va^\top  \mI^{-1} 
        +
        \mI^{-2} \mA \vm \vm^\top \mI^{-2} \mA
        )
        + \circ(\gamma^2)
    \end{align*}
    so that the second term is written as
    \begin{align*}
        & \bigg( \gamma^{-1}T^{-1} + \gamma^0 2T^{-1} + \gamma^1 T^{-1} \bigg)
        \mI_w^{-1} \vm_w \vm_w^\top \mI_w^{-1} \\
        & = 
        \gamma^{-1} \frac{1}{T} (
        \mI^{-1} \vm \vm^\top \mI^{-1}
        )
        +
        \gamma^{0} \frac{2}{T} ( 
        \mI^{-1} \vm \vm^\top \mI^{-1}
        )
        + \\
        & \quad
        \gamma^{1} \frac{1}{T} (
        \mI^{-1} \vm \vm^\top \mI^{-1} +
        \mI^{-1} \va \va^\top  \mI^{-1} +
        \mI^{-2} \mA \vm \vm^\top \mI^{-2} \mA
        )
        + \circ(\gamma)
        \enspace .
    \end{align*}
    The first term of the MSE without the trace is
    \begin{align*}
        & \bigg(\gamma^{-1} T^{-1} + \gamma^0 T^{-1} \bigg) 
        (\mI^{-1}) \\
        & = 
        \bigg(\gamma^{-1} T^{-1} + \gamma^0 T^{-1} \bigg) 
        \bigg(
        \mI^{-1} + \gamma \mI^{-2} \mA + \gamma^2 (\mI^{-1}(\mI^{-1} \mA)^2 - \mI^{-2} \mB) + \circ(\gamma^2)
        \bigg) \\
        & = 
        \gamma^{-1} \frac{1}{T}\mI^{-1} + 
        \gamma^{0}\frac{1}{T}(\mI^{-2} \mA + \mI^{-1}) 
        +
        \gamma^1
        \frac{1}{T}[\mI^{-1}(\mI^{-1} \mA)^2 - \mI^{-2} \mB + \mI^{-2} \mA ]
        + \circ(\gamma) \enspace .
    \end{align*}
    
    Subtracting the second term from the first term and applying the trace, we finally write the MSE:
    \begin{align*}
        & \mathrm{MSE}_{\mathrm{NCE}} 
        \\
        & = 
        \frac{1}{T}
        \mathrm{tr}\bigg(
        \gamma^{-1} \big(
        \mI^{-1} 
        -
        \mI^{-1} \vm \vm^\top \mI^{-1}
        \big)
        + 
        \gamma^{0} \big(
        \mI^{-2} \mA + \mI^{-1}
        - 2\mI^{-1} \vm \vm^\top \mI^{-1}
        \big) 
        \bigg)
        + \circ(\gamma^0) \\
        & = 
        \gamma^{-1} \frac{1}{T}
        \langle \mI_n^{-2}, \Var_{\vd \sim \pd}
        \nabla_{\vbeta} F(\vd, \vbeta^*)
        \rangle
        + 
        \gamma^{0} \frac{1}{T}
        \langle \mI_n^{-2}, 
        (\mA - \vm \vm^\top)
        +
        \Var_{\vd \sim \pd}
        \nabla_{\vbeta} F(\vd, \vbeta^*)
        \rangle
        + \circ(\gamma^0)
    \label{eq:allnoisemse}
    \end{align*}
    Compare this with the MSE obtained with the KL classification loss computed using~\eqref{eq:asympmsegnce}, which is exactly
    \begin{align*}
        \mathrm{MSE}_{\mathrm{KL}} 
        = 
        \gamma^{-1} \frac{1}{T}
        \langle \mI_n^{-2}, \Var_{\vd \sim \pd}
        \nabla_{\vbeta} F(\vd, \vbeta^*)
        \rangle
        + 
        \gamma^{0} \frac{1}{T}
        \langle \mI_n^{-2}, 
        (\mA - \vm \vm^\top)
        \\
        +
        \Var_{\vd \sim \pd}
        \nabla_{\vbeta} F(\vd, \vbeta^*)     
        \rangle
        +
        \gamma^{1} \frac{1}{T}
        \langle \mI_n^{-2}, 
        (\mA - \vm \vm^\top)
        \rangle
        + \circ(\gamma)
    \end{align*}
    In other words, the estimation error (MSE) of NCE with the JS loss matches that of the KL loss in the all noise limit, stopping at order $\gamma^0$. 

    \item Optimize the $\mathrm{MSE}$ w.r.t. $\pn$
    
    To optimize w.r.t. $\pn$, we need only keep the two first orders of the $\mathrm{MSE}_{\mathrm{NCE}}$, which depends on $\pn$ only via the term $\mathrm{tr}(\mI^{-2}\mA) = \int \| \mI^{-1} \nabla_{\vbeta} F(\vx, \vbeta^*) \|^2 \frac{\pd^2}{\pn}(x)d\vx$. This is, up to a multiplicative constant, $\mathcal{D}_{\chi^2}( p_d \bullet w_1, p_n)$
    where the data reweighting function $w_1$ is
    \begin{align*}
        w_1(\vx) 
        = 
        \| \nabla_{\vbeta} F(\vx; \vbeta^*) \|_{\mI^{-2}}
    \end{align*}
    Minimizing this divergence yields
    \begin{equation}
        \pn(\vx) = \|\mI^{-1} \nabla_{\vbeta} F(\vx, \vbeta^*) \| \pd(\vx) /Z
    \end{equation}
    where $Z=\int  \|\mI^{-1} \nabla_{\vbeta} F(\vx, \vbeta^*) \| \pd(\vx)d\vx$ is the normalization constant. This is thus the optimal noise distribution, as a first-order approximation. 
    
    \item Compute the MSE gaps at optimality
    
    Plugging this optimal $\pn$ into the formula of $\mathrm{MSE}_{\mathrm{NCE}}$ and subtracting the Cramer-Rao MSE (which is a lower bound for a normalized model), we get:
    \begin{align*}
        \mathrm{MSE}_{\mathrm{NCE}}(\pn = \pn^{\mathrm{opt}}) - \mathrm{MSE}_{\mathrm{Cramer-Rao}}
        = 
        \frac{1}{T} \bigg( \int \|\mI^{-1} \nabla_{\vbeta} F(., \vbeta^*) \| \pd \bigg)^2 \enspace .
    \end{align*}
    This is interesting to compare with the case where the noise distribution is the data distribution, which gives
    \begin{align*}
        \mathrm{MSE}_{\mathrm{NCE}}(\pn = \pd) - \mathrm{MSE}_{\mathrm{Cramer-Rao}} 
        = 
        \frac{1}{T} \int \|\mI^{-1} \nabla_{\vbeta} F(., \vbeta^*)\|^2 \pd
    \end{align*}

    where the squaring is in a different place.
    In fact, we can compare these two quantities by the Cauchy-Schwartz inequality, or simply the fact that
    \begin{align*}
        \mathrm{MSE}_{\mathrm{NCE}}(\pn = \pd) -
        \mathrm{MSE}_{\mathrm{NCE}}(\pn = \pn^{\mathrm{opt}})
        = 
        \frac{1}{T} \text{Var}_{\vx \sim \pd} \{ \|\mI^{-1} \nabla_{\vbeta} F(\vx, \vbeta^*) \| \}
    \end{align*}
    This implies that the two MSEs, when the noise distribution is either $\pn^{\mathrm{opt}}$ or $\pd$, can only be equal if $ \|\mI^{-1} \vpsi(.)\|$ is constant in the support of $\pd$. This does not seem to be possible for any reasonable distribution.

\end{enumerate}

\noindent \textbf{Proof for noise close to data distribution} We now prove that this  same optimal noise can be obtained in another case, when $\pn \approx \pd$. \\ \\
We consider the limit case where $\frac{\pd}{\pn}(\vx) = 1 + \epsilon(\vx)$ with $|\epsilon(\vx) - 0| < \epsilon_{\mathrm{max}} \quad \forall \vx$.
Note that in order to use Taylor expansions for terms containing $\epsilon(\vx)$ in an integral, we assume for any integrand $h(\vx)$ that $\int h(\vx) \epsilon(\vx) d\vx \approx \eps \int h(\vx) d\vx$, where $\epsilon$ would be a constant.
\begin{enumerate}
    \item Taylor expansion of the reweighting
    \begin{align*}
        w(\vx) 
        & = 
        \frac{\nu \pn(\vx)}{\pd(\vx) + \nu \pn(\vx)}
        = 
        \frac{1}{1 + \frac{1}{\nu} + \frac{\pd}{\pn}(\vx)}
        = 
        \frac{1}{1 + \frac{1}{\nu} + \frac{1}{\nu} \epsilon(\vx)}
        \\
        &
        =
        \frac{\nu}{1 + \nu}\epsilon^0(\vx)
        -
        \frac{\nu}{(1 + \nu)^2}\epsilon^1(\vx)
        +
        \frac{\nu}{(1 + \nu)^3}\epsilon^2(\vx)
        + \circ(\epsilon^2) 
    \end{align*}
    
    \item Evaluating the integrals $\vm_w$, $\mI_w$
    \begin{align*}
        \vm_w 
        & = 
        \int \nabla_{\vbeta} F(\vx, \vbeta^*) 
        w(\vx) \pd(\vx) d\vx \\
        & = 
        \int \nabla_{\vbeta} F(\vx, \vbeta^*)\pd(\vx)
        \bigg( 
        \frac{\nu}{1 + \nu}\epsilon^0(\vx)
        -
        \frac{\nu}{(1 + \nu)^2}\epsilon^1(\vx)
        +
        \frac{\nu}{(1 + \nu)^3}\epsilon^2(\vx)
        + \circ(\epsilon^2)  
        \bigg) 
        d\vx \\     
        & = 
        \frac{\nu}{1 + \nu} \vm 
        - \frac{\nu}{(1 + \nu)^2} \va(\epsilon)
        + \frac{\nu}{(1 + \nu)^3} \vb(\epsilon^2)
        + \circ(\epsilon^3) 
    \end{align*}
    where $\vm$ is the Fisher-score mean, and we use shorthand notations $a$ and $b$ for the remaining integrals:
    \begin{align*}
        \vm
        = 
        \int \nabla_{\vbeta} F(\vx, \vbeta^*)\pd(\vx)
        d\vx
    \end{align*}
    \vspace{-1.5em}
    \begin{alignat*}{2}
        \va(\epsilon)
        = 
        \int \nabla_{\vbeta} F(\vx, \vbeta^*) \pd(\vx) \epsilon(\vx)
        d\vx
        \hspace{2em}
        \vb(\epsilon^2)
        = 
        \int \nabla_{\vbeta} F(\vx, \vbeta^*) \pd(\vx) \epsilon^2(\vx)
        d\vx \enspace .
    \end{alignat*}
    Similarly,
    \begin{align*}
        \mI_w
        & = 
        \int \nabla_{\vbeta} F(\vx, \vbeta^*)\nabla_{\vbeta} F(\vx, \vbeta^*)^\top w(\vx) \pd(\vx) d\vx 
        \\ & = 
        \int \nabla_{\vbeta} F(\vx, \vbeta^*)\nabla_{\vbeta} F(\vx, \vbeta^*)^\top \pd(\vx)
        \bigg( 
        \frac{\nu}{1 + \nu}\epsilon^0(\vx)
        -
        \frac{\nu}{(1 + \nu)^2}\epsilon^1(\vx)
        + 
        \frac{\nu}{(1 + \nu)^3}\epsilon^2(\vx)
        \\
        & + \circ(\epsilon^2)  
        \bigg) 
        d\vx    
        = 
        \frac{\nu}{1 + \nu} \mI
        - \frac{\nu}{(1 + \nu)^2} \mA(\epsilon)
        + \frac{\nu}{(1 + \nu)^3} \mB(\epsilon^2)
        + \circ(\epsilon^3) 
    \end{align*}
    where the Fisher-score covariance (Fisher information) is $\mI$ and we use shorthand notations $\mA$ and $\mB$ for the remaining integrals:
    \begin{align*}
        \mI
        & = 
        \int \nabla_{\vbeta} F(\vx, \vbeta^*)\nabla_{\vbeta} F(\vx, \vbeta^*)^\top \pd(\vx)
        d\vx
        \\
        \mA(\epsilon)
        & = 
        \int \nabla_{\vbeta} F(\vx, \vbeta^*)\nabla_{\vbeta} F(\vx, \vbeta^*)^\top \pd \epsilon(\vx)
        d\vx
        \\
        \mB(\epsilon^2)
        & = 
        \int \nabla_{\vbeta} F(\vx, \vbeta^*)\nabla_{\vbeta} F(\vx, \vbeta^*)^\top \pd \epsilon^2(\vx)
        d\vx \enspace .
    \end{align*}
    \item Taylor expansion of $\mI^{-1}$
    %
    \begin{align*}
        \mI_w^{-1}
        & = 
        \bigg(
        \frac{\nu}{1 + \nu} \mI
        - \frac{\nu}{(1 + \nu)^2} \mA(\epsilon)
        + \frac{\nu}{(1 + \nu)^3} \mB(\epsilon^2)
        + \circ(\epsilon^3)
        \bigg)^{-1} \\
        & = 
        \frac{1 + \nu}{\nu} \mI^{-1} 
        + \frac{1}{\nu} \mI^{-2} \mA(\epsilon) 
        + \frac{\nu}{1 + \nu} \mI^{-2} 
        \big(
        \mI^{-1} \mA^2(\epsilon) - \mB(\epsilon^2)
        \big)
        + \circ(\epsilon^3)
    \end{align*}
    \item Evaluating the $\mathrm{MSE}_{\mathrm{NCE}}$
    \begin{align*}
        & \mI_w^{-1} \vm_w \vm_w^\top \mI_w^{-1}
        \\
        & = 
        \mI^{-1} \vm \vm^\top \mI^{-1}
        +
        \frac{1}{(1 + \nu)^2} \bigg(
        \mI^{-2} \mA(\epsilon) \vm \vm^\top  \mI^{-2} \mA(\epsilon)
        +
        \mI^{-1} \va(\epsilon) \va(\epsilon)^\top \mI^{-1} 
        \bigg)
        + \circ(\epsilon^3)
    \end{align*}
    by plugging in the Taylor expansions of $\mI^{-1}$ and $\vm$ and retaining only terms up to the second order. Finally, the MSE becomes:
    \begin{alignat*}{2}
        &\mathrm{MSE}_{\mathrm{NCE}}(T, \nu, \pn) 
        && = 
        \frac{\nu + 1}{T} \mathrm{tr} (\mI_w^{-1} - \frac{\nu + 1}{\nu} (\mI_w^{-1} \vm_w \vm_w^\top \mI_w^{-1})) 
        \\
        & {}
        && = 
        \mathrm{tr} \bigg(
        \frac{(1 + \nu)^2}{T \nu} (
        \mI^{-1}
        -
        \mI^{-1} \vm_F \vm_F^\top \mI^{-1}
        )
        + 
        \frac{1 + \nu}{T \nu} \mI^{-2} \mA(\epsilon) 
        + 
        \\
        & \frac{1}{T \nu} \big(
        \mI^{-3} \mA^2(\epsilon) 
        -
         \mI^{-2} && \mB(\epsilon^2) 
        - 
        \mI^{-1} \va(\epsilon) \va(\epsilon)^\top \mI^{-1}
        -
        \mI^{-2} \mA(\epsilon) \vm \vm^\top  \mI^{-2} \mA(\epsilon)
        \big)
        \bigg)
        + 
        \circ(\epsilon^3)
    \end{alignat*}
    
    \item Optimize the $\mathrm{MSE}_{\mathrm{NCE}}$ w.r.t. $\pn$
    
    To optimize w.r.t. $\pn$, we need only keep the $\mathrm{MSE}_{\mathrm{NCE}}$ up to order 1, which depends on $\pn$ only via the term
    \begin{align*}
        \mathrm{tr}(\mI^{-2}\mA(\epsilon)) 
        & = 
        \mathrm{tr} \bigg(
        \mI^{-2}   
        \big(
        \int \nabla_{\vbeta} F(\vx, \vbeta^*)\nabla_{\vbeta} F(\vx, \vbeta^*)^\top \frac{\pd^2}{\pn}(x)d\vx
        - 
        \mI
        \big)
        \bigg) \enspace ,
    \end{align*}
    where we unpacked $\pn$ from $\epsilon = \frac{\pd}{\pn} - 1$. Hence, we need to optimize
    \begin{equation}
    J (\pn)= \frac{1}{T} \int \|\mI^{-1} \nabla_{\vbeta} F(\vx, \vbeta^*)\|^2\frac{\pd^2}{\pn}(\vx)d\vx
    \end{equation}
    with respect to $\pn$. This was already done in the all-noise limit $\nu \rightarrow \infty$ and yielded
    \begin{equation}
    \pn(\vx) =   \|\mI^{-1} \nabla_{\vbeta} \log p(x, \vbeta^*) \| \pd(\vx) /Z
    \end{equation}
    where $Z=\int  \|\mI^{-1} \nabla_{\vbeta} F(\vx, \vbeta^*) \| \pd(\vx)d\vx$ is the normalization constant. This is thus the optimal noise distribution, as a first-order approximation. 
\end{enumerate}

\subsection{Limit of all data}
\label{app:ssec:proof_mse_alldata_parametric}

We here prove Conjecture~\ref{th:alldatabestmse}. 
%
 We recall the $\mathrm{MSE}(T, \nu, \pn) = \frac{\nu + 1}{T} \mathrm{tr} (\mI^{-1} - \frac{\nu + 1}{\nu} (\mI^{-1} \vm \vm^\top \mI^{-1}))$.
Given the term up until $\nu^{-1}$ in the MSE, we will use Taylor expansions up to order 2 throughout the proof, in anticipation that the MSE will be expanded until order 1.
Note that in this no noise limit, the assumption made by Gutmann and Hyv\"arinen (2012) that $\pn$ is non-zero whenever $\pd$ is nonzero is not true for this optimal $\pn$, which reduces the rigour of this analysis. (This we denote by heuristic approximation~1.)
\begin{enumerate}
    \item Taylor expansion of the reweighting
    \begin{align*}
        w(\vx)
        = 
        \frac{\nu \pn(\vx)}{\pd(\vx) + \nu \pn(\vx)}
        = 
        \frac{1}{1 + \frac{1}{\nu} \frac{\pd}{\pn}(\vx)}
        = 
        \nu \frac{\pn}{\pd}(\vx) - \nu^2 \frac{\pn^2}{\pd^2}(\vx) + \circ(\nu^2)
    \end{align*}
    
    \item Evaluating the integrals $\vm_w$, $\mI_w$
    
    \begin{align*}
        \vm_w
        & = 
        \int \nabla_{\vbeta} F(\vx, \vbeta^*)\pd(\vx)
        \bigg(1 - D(\vx)\bigg) 
        d\vx 
        = 
        \int \nabla_{\vbeta} F(\vx, \vbeta^*)\pd(\vx)
        \bigg( \nu \frac{\pn}{\pd}(\vx) - 
        \\
        & \nu^2 \frac{\pn^2}{\pd^2}(\vx) + \circ(\nu^2) \bigg) 
        d\vx 
        = 
        \nu \vm_n - \nu^2 \vb + \circ(\nu^2) 
    \end{align*}
    where
    \begin{alignat*}{2}
        \vm_n
        = 
        \int \nabla_{\vbeta} F(\vx, \vbeta^*)\pn(\vx)
        d\vx
        \hspace{2em}
        \vb
        = 
        \int \nabla_{\vbeta} F(\vx, \vbeta^*)\frac{\pn^2}{\pd}(\vx)
        d\vx \enspace .
    \end{alignat*}
    Similarly,
    \begin{align*}
        \mI_w 
        & = 
        \int \nabla_{\vbeta} F(\vx, \vbeta^*)\nabla_{\vbeta} F(\vx, \vbeta^*)^\top 
        w(\vx) \pd(\vx) d\vx 
        = 
        \int \nabla_{\vbeta} F(\vx, \vbeta^*)\nabla_{\vbeta} F(\vx, \vbeta^*)^\top \pd(\vx)
        \bigg( 
        \\
        & \nu \frac{\pn}{\pd}(\vx) - \nu^2 \frac{\pn^2}{\pd^2}(\vx) 
        + \circ(\nu^2) \bigg) 
        d\vx     
        = 
        \nu \mI_n - \nu^2 \mB + \circ(\nu^2) 
    \end{align*}
    where the score covariance evaluated over the noise distribution is $\mI_n$ and we use shorthand notation $\mB$ for the remaining integral:
    \begin{alignat*}{2}
        \mI_n
        & = 
        \int \nabla_{\vbeta} F(\vx, \vbeta^*)\nabla_{\vbeta} F(\vx, \vbeta^*)^\top \pn(\vx)
        d\vx 
        \hspace{0.5em}
        \mB
        & = 
        \int \nabla_{\vbeta} F(\vx, \vbeta^*)\nabla_{\vbeta} F(\vx, \vbeta^*)^\top \frac{\pn^2(\vx)}{\pd(\vx)}
        d\vx 
        .
    \end{alignat*}
    \item Taylor expansion of $\mI_w^{-1}$
    \begin{align*}
        \mI_w^{-1} 
        & = 
        \bigg(
        \nu \mI_n - \nu^2 \mB + \circ(\nu^2) 
        \bigg)^{-1}
        = 
        \bigg(
        \nu\mI_n (\textbf{Id} - \nu \mI_n^{-1} \mB)
        + \circ(\nu^2) 
        \bigg)^{-1} \\
        & = 
        \nu^{-1} \mI_n^{-1}
        \bigg(
        \textbf{Id} + \nu \mI_n^{-1} \mB + \nu^2 (\mI_n^{-1} \mB)^2 + \nu^3(\mI_n^{-1} \mB)^3 + \circ(\nu^3) \bigg)
        + \circ(\nu^2) \\
        & = 
        \nu^{-1} \mI_n^{-1} + \nu^0 \mI_n^{-2} \mB + \nu^1 \mI_n^{-1}(\mI_n^{-2} \mB)^2 + \nu^2 \mI_n^{-1}(\mI_n^{-2} \mB)^3 + \circ(\nu^2)
    \end{align*}
    \item Evaluating the $\mathrm{MSE}$
    \begin{align*}
        & \mI_w^{-1} \vm_w \vm_w^\top \mI_w^{-1}
        = \\
        & \nu^0 (\mI_n^{-1} \vm_n \vm_n^{T} \mI_n^{-1}) 
        +
        \nu^2 (\mI_n^{-1} \vb \vb^{T} \mI_n^{-1} + 
        \mI_n^{-2} \mB \vm_n \vm_n^{T} \mI_n^{-2} \mB)
        + \circ(\nu^2)
    \end{align*}
    by plugging in the Taylor expansions of $\mI_w^{-1}$ and $\vm_w$ and retaining only terms up to the second order. Hence, the second term of the MSE without the trace is
    \begin{align*}
        & ( \nu^{1}T^{-1} + \nu^0 2T^{-1} + \nu^{-1}T^{-1} )
        \mI_w^{-1} \vm_w \vm_w^\top \mI_w^{-1} \\
        & = 
        ( \nu^{1}T^{-1} + \nu^0 2T^{-1} + \nu^{-1}T^{-1} )
        \big(
        \nu^0 (\mI_n^{-1} \vm_n \vm_n^\top \mI_n^{-1}) 
        +
        \\
        & \nu^2 (\mI_n^{-1} \vb \vb^\top \mI_n^{-1} + 
        \mI_n^{-2} \mB \vm_n \vm_n^\top \mI_n^{-2} \mB)
        + \circ(\nu^2)
        \big) \\
        & = 
        \nu^{-1} \frac{1}{T} (\mI_n^{-1} \vm_n \vm_n^\top \mI_n^{-1}) 
        +
        \nu^{0} \frac{1}{T} (2\mI_n^{-1} \vm_n \vm_n^\top \mI_n^{-1})
        + \\
        & \enspace \nu^1 \frac{1}{T} (\mI_n^{-1} \vb_n \vb_n^\top \mI_n^{-1} + 
        \mI_n^{-2} \mB \vm_n \vm_n^\top \mI_n^{-2} \mB + \mI_n^{-1} \vm_n \vm_n^\top \mI_n^{-1})
        + \circ(\nu)
    \end{align*}
    and the first term of the MSE without the trace is
    \begin{align*}
        & ( \nu^{0} T^{-1} + \nu^1 T^{-1} ) 
        \mI_w^{-1} \\
        & = 
        ( \nu^{0} T^{-1} + \nu^1 T^{-1} ) 
        \big(
        \nu^{-1} \mI_n^{-1} + \nu^0 \mI_n^{-2} \mB + \nu^1 \mI_n^{-1}(\mI_n^{-2} \mB)^2 + \nu^2 \mI_n^{-1}(\mI_n^{-2} \mB)^3 + \circ(\nu^2)
        \big) \\
        & = 
        \nu^{-1} \frac{1}{T}\mI_n^{-1} 
        + 
        \nu^{0}\frac{1}{T}(\mI_n^{-2} \mB + \mI_n^{-1}) 
        +
        \nu^1
        \frac{1}{T}[\mI_n^{-1}(\mI_n^{-1} \mB)^2 + \mI_n^{-2} \mB]
        + \circ(\nu) \enspace .
    \end{align*}
    
    Subtracting the second term from the first term and applying the trace, we finally write the MSE:
    \begin{alignat*}{2}
        & \mathrm{MSE} 
        && = \frac{1}{T} \mathrm{tr}(
        \nu^{-1} (\mI_n^{-1} - \mI_n^{-1} \vm_n \vm_n^{T} \mI_n^{-1})
        + 
        \nu^{0} (\mI_n^{-2} \mB + \mI_n^{-1} - 2\mI_n^{-1} \vm_n \vm_n^{T} \mI_n^{-1}) 
        + 
        \nu^1
        [
        \\
        & \mI_n^{-1} ( 
        && \mI_n^{-1} \mB)^2 
        + \mI_n^{-2} \mB  - \mI_n^{-1} \vb_n \vb_n^{T} \mI_n^{-1} - 
        \mI_n^{-2} \mB \vm_n \vm_n^{T} \mI_n^{-2} \mB - \mI_n^{-1} \vm_n \vm_n^{T} \mI_n^{-1}]
        + \circ(\nu) 
        ) .
    \end{alignat*}
    
    Rewriting $\mI_n^{-1} = \mI_n^{-1} \mI_n \mI_n^{-1}$, using the circular invariance of the trace operator, we get:
    \begin{align*}
        \mathrm{MSE}
        = &
        \nu^{-1} \frac{1}{T}
        \langle \mI_n^{-2}, \mI_n - \vm_n \vm_n^\top \rangle
        + 
        \nu^{0}\frac{1}{T}
        \langle \mI_n^{-2}, (\mB - \vm_n \vm_n^\top) + (\mI_n - \vm_n \vm_n^\top) \rangle
        \\
        & \nu^1 \frac{1}{T} \langle 
        \mI_n^{-2}, (\mI_n - \vm_n \vm_n^\top)
        + \mB \mI_n^{-1} \mB - \vb_n \vb_n^\top - \mB \vm_n \vm_n^\top \mI_n^{-2} \mB 
        \rangle
        + \circ(\nu) \\
        = & 
        \nu^{-1} \frac{1}{T}
        \langle \mI_n^{-2}, \Var_{\vn \sim \pn}
        \nabla_{\vbeta} F(\vn, \vbeta^*)
        \rangle
        + 
        \nu^{0}\frac{1}{T}
        \langle \mI_n^{-2}, (\mB - \vm_n \vm_n^\top) 
        + 
        \\
        & \Var_{\vn \sim \pn}
        \nabla_{\vbeta} F(\vn, \vbeta^*)
        \rangle
        + \nu^1 \frac{1}{T} \langle 
        \mI_n^{-2}, \Var_{\vn \sim \pn}
        \nabla_{\vbeta} F(\vn, \vbeta^*)
        + \mB \mI_n^{-1} \mB - \vb_n \vb_n^\top 
        \\
        & - \mB \vm_n \vm_n^\top \mI_n^{-2} \mB 
        \rangle
        + \circ(\nu) \\
        & = 
        \nu^{-1} \frac{1}{T}
        \langle \mI_n^{-2}, \Var_{\vn \sim \pn}
        \nabla_{\vbeta} F(\vn, \vbeta^*)
        \rangle
        + 
        \nu^{0}\frac{1}{T}
        \langle \mI_n^{-2}, \mB - \vm_n \vm_n^\top + 
        \\
        & \hspace{1em} \Var_{\vn \sim \pn}
        \nabla_{\vbeta} F(\vn, \vbeta^*)
        \rangle
        + \circ(1) .
    \numberthis
    \label{eq:alldatamse}
    \end{align*} 
    Compare this with $\mathrm{MSE}_{\mathrm{revKL}}$ computed using~\eqref{eq:asympmsegnce}, which is exactly
    \begin{align*}
        \mathrm{MSE}_{\mathrm{revKL}}
        & = 
        \nu^{-1} \frac{1}{T}
        \langle \mI_n^{-2}, \Var_{\vn \sim \pn}
        \nabla_{\vbeta} F(\vn, \vbeta^*)
        \rangle
        + 
        \nu^{0}\frac{1}{T}
        \langle \mI_n^{-2}, (\mB - \vm_n \vm_n^\top) 
        \\
        & + \Var_{\vn \sim \pn}
        \nabla_{\vbeta} F(\vn, \vbeta^*)
        \rangle
        + \nu^1 \frac{1}{T} \langle 
        \mI_n^{-2}, \Var_{\vn \sim \pn}
        \nabla_{\vbeta} F(\vn, \vbeta^*)
        \rangle
        \enspace .
    \end{align*}
    In other words, the estimation error (MSE) of NCE with the JS loss matches that of the revKL loss in the all data limit, stopping at order $\nu^0$. 

    \item Optimize the $\mathrm{MSE}_{\mathrm{NCE}}$ w.r.t. $\pn$
    
    Looking at the above MSE, the dominant term of order $\nu^{-1}$ is given by \\
    $\langle \mI_n^{-2}, \Var_{\vn \sim \pn} \nabla_{\vbeta} F(\vn, \vbeta^*) \rangle \geq 0$. It is minimized when it is $0$, that is, when $\nabla_{\vbeta} F(., \vbeta^*)$ is constant in the support of $\pn$. Typically this means that $\pn$ is concentrated on a set of zero measure. In the 1D case, such case is typically the Dirac delta $\pn = \delta_z$, or a distribution with two deltas in case of symmetrical $\vpsi$.
    
    We can plug this in the terms of the next order $\nu^0$, which remain to be minimized: 
    \begin{align*}
        \langle \mI_n^{-2}, \mB + \mI_n - 2 \vm_n \vm_n^{T} \rangle
        & = 
        \langle \mI_n^{-2}, \mB - \mI_n + 2\mI_n - 2 \vm_n \vm_n^{T} \rangle \\
        & =
        \langle \mI_n^{-2}, \mB - \mI_n + 2\Var_{\vn \sim \pn}
        \nabla_{\vbeta} F(\vn, \vbeta^*)         \rangle 
        \\ 
        & =
        \langle \mI_n^{-2}, \mB - \mI_n \rangle
    \end{align*} 
    given we chose $\pn$ so that the variance is 0.
    
    The integrands of $\mB$ and $\mI_n$ respectively involve $\pn^2$ and $\pn$. Because $\pn$ is concentrated on a set of zero measure (Dirac-like), the term in $\mB$ dominates the term in $\mI_n$. This is because if we consider the $\pn$ as the limit of a sequence of some proper pdf's, the value of the pdf gets infinite in the support of that pdf in the limit, and thus $\pn^2$ is infinitely larger than $\pn$. Hence we are left with $\langle \mI_n^{-2}, \mB \rangle$.

    The matrix $\mB$ 
    simplifies to evaluating the $\nabla_{\vbeta} F(\vx, \vbeta^*)\nabla_{\vbeta} F(\vx, \vbeta^*)^\top/\pd(\vx)$ in the support of $\pn$. Since we know that $\nabla_{\vbeta} F(\vx, \vbeta^*)$ is constant in that set, the main question is whether $\pd$ is constant in that set as well. Here, we heuristically assume that it is; this is intuitively appealing in many cases, if not necessarily true. (This we denote by heuristic approximation~2.)
    
    Thus, we have
    \begin{align*}
    \mB 
    &=
    \int \nabla_{\vbeta} F(\vx, \vbeta^*)\nabla_{\vbeta} F(\vx, \vbeta^*)^\top \frac{\delta_z^2}{\pd}(\vx)
    d\vx 
    \approx 
    c\; 
    \nabla_{\vbeta} F(\vz, \vbeta^*)
    \nabla_{\vbeta} F(\vz, \vbeta^*)^\top 
    \frac{1}{\pd(\vz)}
    \end{align*}
    for some constant $c$ taking into account the effect of squaring of $\pn$ (it is ultimately infinite, but the reasoning is still valid in any sequence going to the limit.)

    As for $\mI_n^{-2}$, it is written as
    \begin{align*}
        \mI_n^{-2}
        =
        \left(
        \int 
        \nabla_{\vbeta} F(\vx, \vbeta^*)
        \nabla_{\vbeta} F(\vx, \vbeta^*)^\top 
        \delta_z(\vx) d\vx
        \right)^{-2}
        =
        \left(
        \nabla_{\vbeta} F(\vz, \vbeta^*)
        \nabla_{\vbeta} F(\vz, \vbeta^*)^\top 
        \right)^{-2}
        \enspace .
    \end{align*}
    Because $\pn$ is a Dirac, computing $\mI_n^{-2}$  requires inverting a rank 1 matrix which is ill-posed in dimensions bigger than 1. We therefore make assumption~3: we suppose the parameter being estimated is a scalar, thus $\psi$ is a scalar-valued function.
    We further obtain
    \begin{align*}
        \langle I_n^{-2}, B \rangle 
        & =
        c \frac{1}{\pd(\vz) 
        (\frac{d}{d \beta} F(\vz, \beta^*))^2}
        =
        \int 
        \big( 
        \frac{d}{d \beta} F(\vz, \beta^*) \big)^{-2} 
        \frac{\delta_z^2(\vx)}{\pd(\vx)} d\vx
    \enspace .
    \numberthis
    \label{eq:alldatamsedominant}
    \end{align*}
    Up to a multiplicative constant, we recognize this as
    \begin{align*}
        \mathcal{D}_{\chi^2}(p_n,  p_d \bullet w_2)
    \end{align*}
    where the noise distribution $\pn$ and data reweighting function $w_2$ are given by
    \begin{alignat*}{2}
        \pn(\vx) 
        &= 
        \delta_z(\vx)   
        \hspace{2em}
        w_2(\vx) 
        &= 
        \big(
        \frac{d}{d \beta} F(\vx, \beta^*)
        \big)^2
    \end{alignat*}

Those points $z$ obtained by the above condition are the best candidates for $\pn$ to concentrate its mass on.

We arrived this result by making two heuristic approximations and an assumption as explained above; we hope to be able to remove some of them in future work.

\end{enumerate}

\subsection{Practical examples for the optimal noise distribution}
\label{sec:appendix:estim_error_practical}

\paragraph{Gaussian Mean} 
Consider a one-dimensional Gaussian data distribution, with mean zero and unit variance. It is here parameterized by its mean and, for an unnormalized model, also by its log-normalization as a free parameter. We can write the following distribution models, respectively the energy, the normalized distribution and the one-parameter extended unnormalized distribution,
\begin{alignat*}{3}
    \tilde{p}(x; \vtheta) 
    =
    \exp \big( \frac{-(x - \theta)^2}{2} \big)
    \hspace{0.5em}
    p(x, \vtheta)
    =
    \frac{1}{\sqrt{2 \pi}}
    \exp \big( \frac{-(x - \theta)^2}{2} \big)
    \hspace{0.5em}
    p(x, \vTheta)
    =
    \exp \big( \frac{-(x - \theta)^2}{2} -c \big)
\end{alignat*}
where $\theta^* = 0$.
We then obtain their respective versions of the Fisher score
\begin{align*}
    \nabla_{\theta} \log \tilde{p}(x, \vtheta^*)
    & = 
    x - \theta^*
    = x
    \\
    \nabla_{\theta} \log p(x, \vtheta^*)
    & = 
    x - \theta^*
    = x
    \\
    \nabla_{\vTheta} \log p(x, \vTheta^*)
    & = 
    [\nabla_{\theta} \log \tilde{p}(x, \vtheta^*), -1]^\top
    = [x, -1]^\top
\end{align*}
Based on these, we can compute the following integrals which appear in the estimation error
\begin{align*}
    m(\theta^*) 
    & = 
    \int 
    \nabla_{\theta} \log \tilde{p}(x, \vtheta^*)
    \pd(x)
    = \E_{\pd}[x] = 0
    \\
    I(\theta^*) 
    & = 
    \int 
    \nabla_{\theta} \log \tilde{p}(x, \vtheta^*)
    \nabla_{\theta} \log \tilde{p}(x, \vtheta^*)^\top
    \pd(x)
    = \E_{\pd}[x^2] = 1
    \\
    \Ifisher(\theta^*) 
    & =
    \int 
    \nabla_{\theta} \log p(x, \vtheta^*)
    \nabla_{\theta} \log p(x, \vtheta^*)^\top
    \pd(x)
    = \E_{\pd}[x^2] = 1
    \\
    \mI(\vTheta^*)^{-1}
    &= 
    \big(
    \int 
    \nabla_{\vTheta} \log p(\vx, \vTheta^*) 
    \nabla_{\vTheta} \log p(\vx, \vTheta^*)^\top
    \pd(x)
    \big)^{-1}
    = \mathrm{Id}
\end{align*}
The optimal noise distribution in the all-noise limit for the \textbf{normalized} model is
\begin{align*}
    \pn(x) 
    & \propto 
    \pd(x) \|  \Ifisher^{-1}(\vtheta^*) \nabla_{\vtheta} \log p(x, \vtheta^*) \|
    \propto 
    \pd(x) |x|
    \enspace .
\end{align*}
The optimal noise distribution in the all-noise limit for the \textbf{unnormalized} model is
\begin{align*}
    \pn(x) 
    & \propto 
    \pd(x) \|  \mI(\vTheta^*)^{-1} \nabla_{\vTheta} \log p(x, \vTheta^*) \|
    \propto 
    \pd(x) \sqrt{x^2 + 1}
    \enspace .
\end{align*}
These two noise distributions are shown in Figures    \ref{fig:optimalnoisemean} and     \ref{fig:optimalnoisemean_ebm}, respectively.

\paragraph{Gaussian Variance} 
Consider a one-dimensional Gaussian data distribution, with mean zero and unit variance. It is here parameterized by its log variance and, for an unnormalized model, also by its log-normalization as a free parameter. We can write the following distribution models, respectively the energy, the normalized distribution and the one-parameter extended unnormalized distribution,
\begin{alignat*}{3}
    \tilde{p}(x; \vtheta) 
    =
    \exp \big( -\frac{1}{2} \frac{x^2}{\theta} \big)
    \hspace{2em}
    p(x, \vtheta)
    =
    \frac{1}{\sqrt{2 \pi \theta}}
    \exp \big( -\frac{1}{2} \frac{x^2}{\theta} \big)
    \hspace{2em}
    p(x, \vTheta)
    =
    \exp \big( -\frac{1}{2} \frac{x^2}{\theta} - c \big)
\end{alignat*}
where $\theta^* = 1$.
We then obtain their respective versions of the Fisher score
\begin{align*}
    \nabla_{\theta} \log \tilde{p}(x, \vtheta^*)
    & = 
    \frac{1}{2} \frac{x^2}{{\theta^*}^2}
    = \frac{1}{2} x^2
    \\
    \nabla_{\theta} \log p(x, \vtheta^*)
    & = 
    \nabla_{\theta} \log \tilde{p}(x, \vtheta^*)
    -
    \nabla_{\theta} \log Z(\theta^*)
    = \frac{1}{2} x^2 - \frac{1}{2}
    \\
    \nabla_{\vTheta} \log p(x, \vTheta^*)
    & = 
    [\nabla_{\theta} \log \tilde{p}(x, \vtheta^*), -1]^\top
    = [\frac{1}{2} x^2, -1]^\top
\end{align*}
Based on these, we can compute the following integrals which appear in the estimation error
\begin{align*}
    m(\theta^*) 
    & = 
    \int 
    \nabla_{\theta} \log \tilde{p}(x, \vtheta^*)
    \pd(x)
    = \E_{\pd}[\frac{1}{2} x^2] = \frac{1}{2}
    \\
    I(\theta^*) 
    & = 
    \int 
    \nabla_{\theta} \log \tilde{p}(x, \vtheta^*)
    \nabla_{\theta} \log \tilde{p}(x, \vtheta^*)^\top
    \pd(x)
    = \E_{\pd}[\frac{1}{4} x^4] = \frac{3}{4}
    \\
    \Ifisher(\theta^*) 
    & =
    \int 
    \nabla_{\theta} \log p(x, \vtheta^*)
    \nabla_{\theta} \log p(x, \vtheta^*)^\top
    \pd(x)
    = \E_{\pd}[\frac{1}{4}(x^2 - 1)^2]
    \\
    & = 
    \frac{1}{4}
    \big(
    \E_{\pd}[x^4] - 2 \E_{\pd}[x^2] + \E_{\pd}[1]
    \big)
    = \frac{1}{4} (3 - 2 + 1) = \frac{1}{2}
    \\
    \mI(\vTheta)^{-1}
    &= 
    \big(
    \int 
    \nabla_{\vTheta} \log p(\vx, \vTheta^*) 
    \nabla_{\vTheta} \log p(\vx, \vTheta^*)^\top
    \pd(x)
    \big)^{-1}
    =
    \bigg[
    \begin{array}{c|c}
        2
        & 1 \\
        \hline
        1
        & 
        \frac{3}{2}
    \end{array}
    \bigg]
\end{align*}
The optimal noise distribution in the all-noise limit for the \textbf{normalized} model is
\begin{align*}
    \pn(x) 
    & \propto 
    \pd(x) \|  \Ifisher^{-1}(\vtheta^*) \nabla_{\vtheta} \log p(x, \vtheta^*) \|
    \propto 
    \pd(x) 
    |x^2 - 1|
    \enspace .
\end{align*}
The optimal noise distribution in the all-noise limit for the \textbf{unnormalized} model is
\begin{align*}
    \pn(x) 
    & \propto 
    \pd(x) \|  \mI(\vTheta^*)^{-1} \nabla_{\vTheta} \log p(x, \vTheta^*) \|
    \propto 
    \pd(x) 
    \sqrt{
    \big( x^2 - 1 \big)^2
    +
    \big( \frac{x^2}{2} - \frac{3}{2} \big)^2
    }
    \enspace .
\end{align*}
These two noise distributions are shown in Figures    \ref{fig:optimalnoisevar} and     \ref{fig:optimalnoisevar_ebm}, respectively.
\omar{Overall, this subsection demonstrates how we can obtain the optimal noise distribution (in the all-noise limit) for simple models where the fisher score vector and its variance can be computed.}

\section{Finding the optimal classification imbalance}
\label{app:sec:gnce_optimal_noise_proportion}

\paragraph{}
So far, we have found that the optimal classification loss (indexed by $\phi$) is the logistic loss, and a suboptimal but acceptable noise distribution is $\pn = \pd$. In this setting, we can now compute the optimal noise proportion $\nu$. 

We wish to minimize the MSE 
\begin{align*}
    \mathrm{MSE}_{\mathrm{NCE}}(T, \nu, \pn) =
    \frac{\nu + 1}{T} \mathrm{tr}
    (
        \mI_w^{-1} & - \frac{\nu + 1}{\nu} (\mI_w^{-1} \vm_w \vm_w^\top \mI_w^{-1})
    )
    \enspace .
\end{align*}
Given that $\pn = \pd$, the reweighting simplifies to 
\begin{align*}
    w(\vx)
    =
    \frac{\nu \pn}{\pd + \nu \pn}(\vx)
    =
    \frac{\nu}{1 + \nu} 
\end{align*}
and the integrals involved become
\begin{align*}
    \vm_w
    & = 
    \int \nabla_{\vbeta} F(\vx, \vbeta^*) w(\vx) \pd(\vx) d\vx
    =
    \frac{\nu}{1 + \nu} \vm
    \\
    \mI_w
    & = 
    \int \nabla_{\vbeta} F(\vx, \vbeta^*)\nabla_{\vbeta} F(\vx, \vbeta^*)^\top w(\vx) \pd(\vx) d\vx
    = 
    \frac{\nu}{1 + \nu} \mI 
    \enspace .
\end{align*}
The MSE thus reduces to
\begin{align*}
    \mathrm{MSE}(\nu) 
    & =
    \frac{(\nu + 1)^2}{\nu T} \mathrm{tr}(
    \mI^{-1} - \mI^{-1} \vm \vm^\top \mI^{-1}
    )    
    \propto
    \frac{(\nu + 1)^2}{\nu} \enspace .
\end{align*}
The derivative with respect to $\nu$ is proportional to $\frac{1}{\nu^2} - 1$ and is null when $\nu=1$ so when the noise proportion is $50\%$. 

\paragraph{}
Furthermore, when the model $p_{\vbeta}$ is normalized, we can compare the $\mathrm{MSE}$ achieved by NCE (using $T_d$ data samples and $T_n$ noise samples) with the $\mathrm{MSE}$ achieved by MLE (using $T_d$ data samples):
\begin{align*}
    \frac{\mathrm{MSE}_{\mathrm{NCE}}(T, \nu, \pn)}{\mathrm{MSE}_{\mathrm{MLE}}(T_d)}
    & =
    \frac{
    \frac{(\nu + 1)^2}{\nu T} \mathrm{tr}(\Ifisher^{-1} )
    }
    {
    \frac{1}{T_d} \mathrm{tr}(\Ifisher^{-1} )
    }
    =    
    \frac{
    \frac{(\nu + 1)^2}{\nu T} \mathrm{tr}(\Ifisher^{-1} )
    }
    {
    \frac{\nu + 1}{T} \mathrm{tr}(\Ifisher^{-1} )
    }
    =    
    1
    +
    \frac{1}{\nu}
\end{align*}
which is known from~\citep{gutmann2012nce,pihlaja2010nce}.

\end{document}